\documentclass{article}

\usepackage{PRIMEarxiv}
\usepackage{float}
\usepackage{graphicx}
\usepackage{subcaption}
\usepackage[utf8]{inputenc} 
\usepackage[T1]{fontenc}    
\usepackage{hyperref}       
\usepackage{url}            
\usepackage{booktabs}       
\usepackage{amsfonts}       
\usepackage{nicefrac}       
\usepackage{microtype}      
\usepackage{lipsum}
\usepackage{graphicx}
\graphicspath{{media/}}     
\usepackage{makecell}  
\usepackage{multirow}
\usepackage{PRIMEarxiv}
\usepackage{threeparttable}
\usepackage[utf8]{inputenc} 
\usepackage[T1]{fontenc}    
\usepackage{hyperref}       
\usepackage{url}            
\usepackage{booktabs}       
\usepackage{amsfonts}       
\usepackage{nicefrac}       
\usepackage{microtype}      
\usepackage{lipsum}
\usepackage{fancyhdr}       
\usepackage{graphicx}       
\graphicspath{{media/}}     
\usepackage[numbers]{natbib}
\usepackage{hyperref}
\usepackage{amsmath} 
\usepackage{makecell}
\usepackage{enumitem}
\usepackage{xcolor}
\usepackage{mdframed}
\usepackage{titlesec} 
\definecolor{promptbg}{RGB}{245,249,255}
\usepackage{multicol} 

\title{CAMB: A comprehensive industrial LLM benchmark on civil aviation maintenance
}

\author{
  Feng Zhang, Chengjie Pang, Yuehan Zhang, Chenyu Luo \\
  \texttt{
  zhangfeng-dc@360.cn,
  longguangfei@gmail.com,
  yhzhang0725@gatech.edu,
  luochenyu@stu.zuel.edu.cn
  } \\
}

\begin{document}
\maketitle

\begin{abstract}
Civil aviation maintenance is a domain characterized by stringent industry standards. Within this field, maintenance procedures and troubleshooting represent critical, knowledge-intensive tasks that require sophisticated reasoning.    To address the lack of specialized evaluation tools for large language models (LLMs) in this vertical, we propose and develop an industrial-grade benchmark specifically designed for civil aviation maintenance.  This benchmark serves a dual purpose:
It provides a standardized tool to measure LLM capabilities within civil aviation maintenance, identifying specific gaps in domain knowledge and complex reasoning. By pinpointing these deficiencies, the benchmark establishes a foundation for targeted improvement efforts (e.g., domain-specific fine-tuning, RAG optimization, or specialized prompt engineering), ultimately facilitating progress toward more intelligent solutions within civil aviation maintenance. Our work addresses a significant gap in the current LLM evaluation, which primarily focuses on mathematical and coding reasoning tasks.  In addition, given that Retrieval-Augmented Generation (RAG) systems are currently the dominant solutions in practical applications , we leverage this benchmark to evaluate existing well-known vector embedding models and LLMs for civil aviation maintenance scenarios. Through experimental exploration and analysis, we demonstrate the effectiveness of our benchmark in assessing model performance within this domain, and we open-source
\thanks{\textit{\underline{this evaluation benchmark and code}} to foster further research and development:
\textbf{https://github.com/CamBenchmark/cambenchmark}}
\end{abstract}
\bibliographystyle{plainnat}



\section{Introduction}
During the past century, civil aviation has evolved into a critical global transportation system, characterized by operational complexity and stringent safety requirements \cite{pcj_add_1}. Despite the maturation of sensor technologies for aircraft inspection, maintenance operations continue to incur substantial labor, temporal, and economic costs \cite{pcj_add_2, pcj_add_3}. Recent advances in Large Language Models (LLMs) have catalyzed multidimensional applications within civil aviation maintenance\cite{pcj_add_4, pcj_add_6}, including civil aviation assembly, safety inspection \cite{pcj_add_5}, civil aviation data analysis \cite{pcj_add_7} to the understanding of complex airworthiness standards and regulations \cite{pcj_add_8}. Mature academic frameworks in higher/vocational education provide standardized curricula for aviation maintenance, supported by authoritative teaching outlines (core content in Fig. 1, see Appendix).  

\begin{figure}
  \centering
  \includegraphics[width=0.8\textwidth]{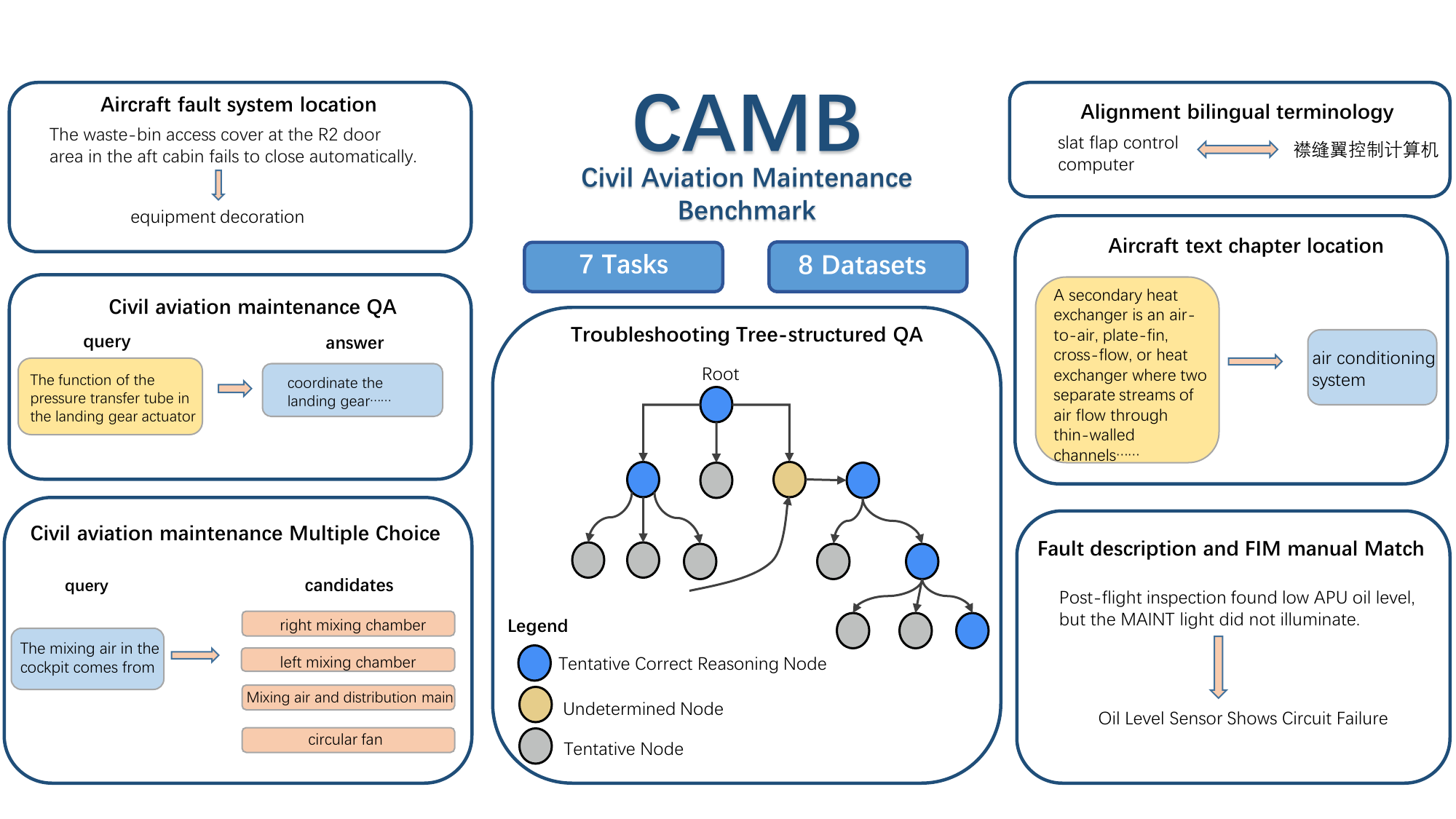}
  \caption{An overview of tasks and datasets in CAMB.}
  \label{fig:fig1}
\end{figure}

Based on the teaching syllabus and our understanding of the civil aviation maintenance field, we integrate industry data to establish a multidimensional evaluation benchmark that features: tasks encompassing fault description, system localization, failure tracing, manual application, and maintenance recommendation; knowledge across aerodynamics, electromechanical control, materials science, and communication systems;  model capability evaluation emphasizing reasoning and engineering decision-making.  This comprehensive benchmark can be used for both the evaluation of embedding models and generalized LLMs.

Recent years have witnessed a surge in specialized benchmarks for LLMs, driving advances in model intelligence and enabling wider application in various fields. This is reflected in general engineering through the creation of standardized evaluation protocols and datasets \cite{pcj_add_9}. In the field of civil aviation, existing benchmarks address critical areas such as safety\cite{pcj_add_11,pcj_add_12} and aviation language understanding\cite{pcj_add_10}. In this work, we present a new evaluation framework designed for LLMs that involves complex knowledge and deep reasoning in civil aviation maintenance. Our key contributions include:

1. An LLM benchmark is proposed and constructed in the field of civil aviation maintenance. The specific evaluation tasks are shown in Table 1.

2. We evaluated state-of-the-art embedding models and LLMs using this benchmark, yielding the following key findings:
\begin{itemize}
\item Current embedding models prioritize semantic similarity but exhibit limited factual knowledge retrieval accuracy for civil aviation maintenance;
\item LLMs achieve multiple-choice task accuracy between 60\%-70\% on our benchmark. In reasoning-intensive tasks, knowledge deficiencies and conceptual ambiguities cause the thinking mode to exhibit certain Test-Time Scaling Law properties, yet fail to significantly surpass the non-thinking mode's instant answering capability;
\item Our civil aviation maintenance evaluation benchmark is effective to distinguish the performance of different embedding models and LLMs.
\end{itemize}

3. Observations from the experimental results and further explorations are from two directions: knowledge embedding and deep reasoning:
\begin{itemize}
\item We conduct exploratory domain adaptation in the embedding model, aiming to improve the knowledge retrieval accuracy for civil aviation maintenance;
\item Based on retrieval-augmented generation systems, using the knowledge retrieval of the embedding model, we aim to improve the question-answering ability of LLMs in civil aviation maintenance.
\end{itemize}

\begin{table}[htbp]
  \centering
  \caption{Benchmark Tasks}
  \begin{tabular}{lll}
   \toprule
    Model Type    & Embedding(Vector)  & LLM(QA) \\
    \midrule
    Alignment bilingual terminology & Recall  & Translate     \\
    Aircraft fault system location  & Rerank & Classification      \\
    Aircraft text chapter location   & Clustering & Classification      \\
    Civil aviation maintenance Multiple Choice    & Rerank       & Multiple Choice QA  \\
    Fault description and FIM manual Match   & Recall       & Pair Classification  \\
    Civil aviation maintenance QA           & Retrieval                          & open QA  \\  
    Fault tree-structured QA   & Rerank     & open QA  \\
    \bottomrule
  \end{tabular}
  \label{tab:horizontal_center}
\end{table}
\section{Related Work}
The application of Large Language Models (LLMs) in industrial systems is an emerging area with significant potential.  \citet{pcj2} explore the critical challenges that LLMs face in real-world industrial settings, including ensuring sufficient accuracy, adherence to industry standards, understanding non-local relationships between entities, understanding industrial process design and product architecture. \citet{lcy1} present a system that integrates PRO-AID for diagnostic tasks and an output-restricted  LLM (mitigating hallucinations) for explanations, validated by deployment in a real factory environment. \citet{lcy2}apply LLMs to PHM fault diagnosis by converting tabular data into text, while investigating the influence of different training data on model's performance. \citet{pcj6}propose a joint KG-LLM system for aviation assembly fault diagnosis, implementing a RAG-based solution that integrates entity retrieval, graph search, and vector retrieval techniques. HybridRAG\cite{zyh4} is a troubleshooting framework that integrates knowledge graphs, multi-dimensional retieval strategy and LLMs to enhance aircraft fault diagnosis efficiency. In addition, the paper presents a planner agent using a plan-execute-reflect (PER) framework to decompose complex tasks, execute steps, and refine strategies for improved reasoning. \citet{pcj5}generate hierarchical troubleshooting tree by first generating a component list, then building branches for each component node. 

Building on diagnostic applications, LLMs are also being actively explored for automated fault localization. \citet{pcj1} demonstrate the use of LLMs for debugging code, identifying fault locations, providing rational explanations, and offering repair suggestions, leading to improvements over existing baselines. \citet{pcj3} introduce a multi-agent framework for code fault localization and repair, showing enhanced performance on the Defects4J benchmark. FD-LLM \cite{zyh2} is a framework specifically designed for fault diagnosis. This framework addresses the challenge of incorporating non-textual data, such as vibration signals by using two methods to encode vibration signals: one converts them into string-based text representations, and the other extracts statistical features from both time and frequency domains. Addressing the challenge of noisy entities in unsupervised entity alignment for aviation fault knowledge graphs, CELLMEA \cite{lcy3} leverages LLMs to inject world knowledge, combining multi-view semantic embeddings, adaptive method for mixing hard negative samples, and incremental consistency regularization technique.

As LLM capabilities in industrial and fault diagnosis contexts advance, the development of robust benchmarks becomes crucial for reliable evaluation.  \citet{zyh3} offer a critical look at the current state of LLM evaluation, highlighting significant challenges in achieving reproducibility, reliability, and robustness. Key concerns include performance variations caused by decoding parameters, potential benchmark contamination and the lack of diversity and coverage in benchmarks. 
ExCyTIn-Bench \cite{zyh1} is the first benchmark specifically designed to assess LLMs in cyber threat investigation. Its exclusion from public training datasets effectively mitigates test-set leakage concerns and thereby demonstrates genuine progress in LLM capabilities.
\citet{pcj4} create a multimodal benchmark in aviation area, which utilized GPT-4o and Claude 3.5 Sonnet with Retrieval-Augmented Generation (RAG), concluded that the current performance of these models is insufficient for the demands of the aerospace industry.  AeroEngQA\cite{lcy4} is a low volume, high quality benchmark aircraft design QA dataset used to evaluate three classes of LLM-based models: zero-shot prompting, in-context prompting, and RAG approaches. \citet{lcy5} introduce evaluation metrics tailored for LLMs in aerospace manufacturing to assess answer accuracy through performance analysis on domain-specific knowledge questions. And as an investigation of LLMs in addressing mechanical engineering problems, \citet{lcy6} demonstrate GPT-4 outperforms other models and human groups in most mechanics domains (excluding continuum mechanics), with performance highly dependent on prompt strategies.

\section{Aircraft maintenance embedded model evaluation benchmark}

\begin{figure}
\centering
\includegraphics[width=170mm, height=90mm]{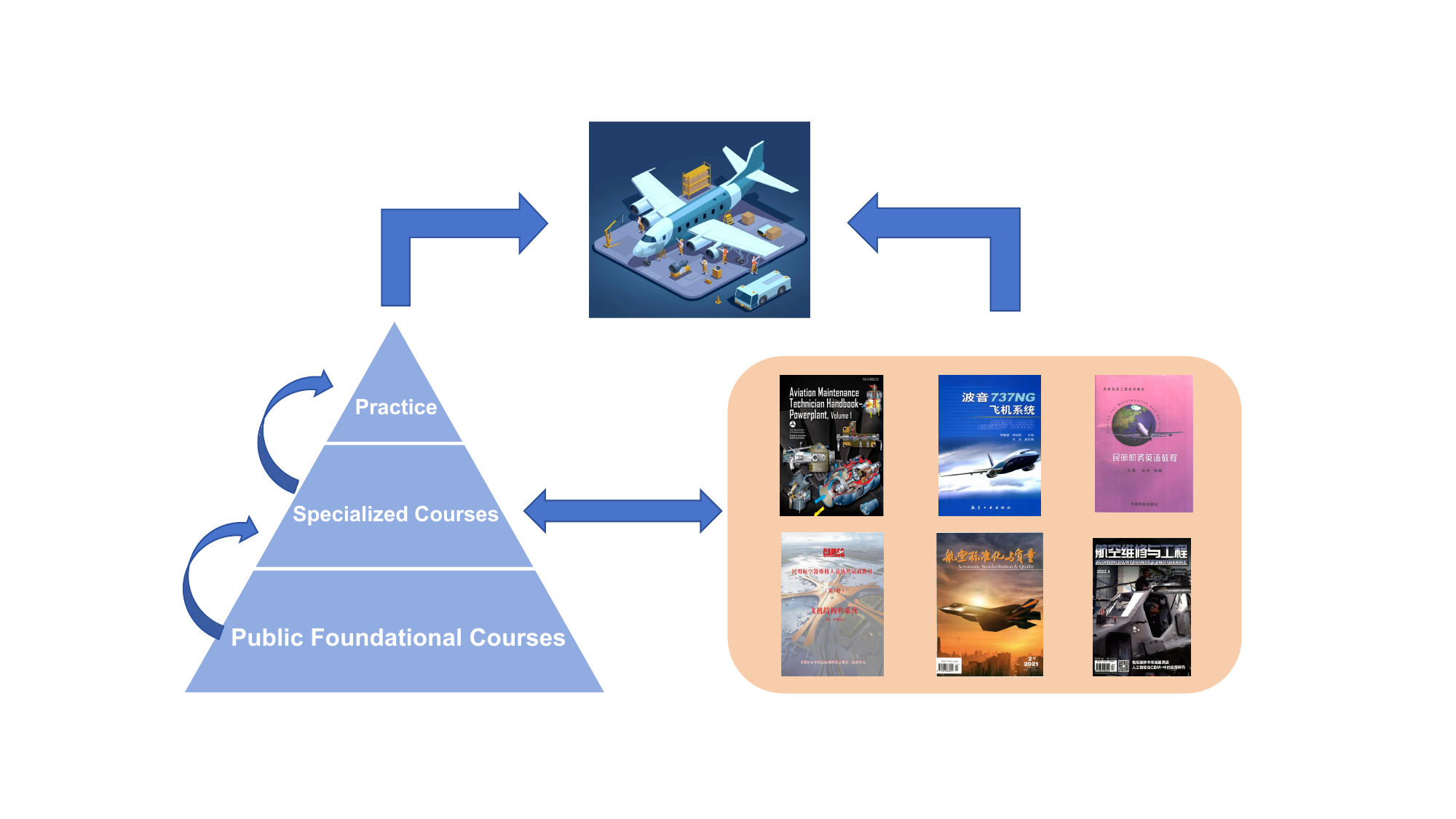}
\caption{Overview of TC-Net structure.}
\label{fig:fig2}
\end{figure}

\subsection{Data Construction}
We collected data from civil aviation maintenance textbooks, bilingual aligned corpora, FIM and TSM manuals, typical fault cases, related articles, exam question-and-answer questions, and exam multiple-choice questions. The maintenance textbooks cover ATA, Boeing 737NG, FAA, and other knowledge about aircraft maintenance principles. The bilingual aligned corpora include granular materials from words to sentences. The FIM and TSM manuals contain information such as fault descriptions and fault entries. The fault cases include real fault phenomena and troubleshooting processes. The exam questions include both question-and-answer and multiple-choice questions for different aircraft types and courses.

\textbf{Chinese-English aligned dataset}:We divided the Chinese and English articles from the textbook 'Aviation Maintenance English' into paragraphs, aligned them using regular expressions, and then reviewed them manually to form aligned Chinese and English paragraphs. Additionally, we translated 100 high-quality fault issues from the FIM manual into Chinese using QWEN3-32B, forming aligned Chinese and English sentences. Finally, we combined these two sets of aligned data with the aviation maintenance bilingual corpus's words and sentences to create an aligned Chinese and English dataset for the Alignment bilingual terminology task.

\textbf{Cluster dataset}:We selected chapters 21 to 30 from ATA, semantically segmented each chapter's content, defined chunksize as 300, organized the segmented text along with its chapter titles into a data format of (text, chapter title), forming a clustering dataset. This dataset is divided into 10 categories for Aircraft text chapter localization task.

\textbf{Categorize dataset}:We manually selected 100 fault issues from the FIM and TSM manuals as well as typical failure cases. We then obtained the first-level systems corresponding to these fault issues from the manuals and cases. The first-level systems are divided into 27 categories. Subsequently, we organized the fault issues and first-level systems into a classification dataset in the format of (fault, first-level system) for Aircraft fault system localization task.

\textbf{Sentence pairing dataset}:We selected fault issues and their detailed descriptions from the FIM manual to form text pairs. Additionally, we chose fault issues and their corresponding entry names from Chapter 49 of the FIM manual (entries refer to detailed reports of aircraft faults used for maintenance analysis and safety improvements) to form another set of text pairs. Then, we combined these two sets of text pairs to create positive sample text pairs. For negative sample text pairs, we used BGE-large-v1.5, gte-Qwen2-1.5B-instruct, gte-Qwen2-7B-instruct, Qwen3-Embedding-4B, and Qwen3-Embedding-8B to evenly divide the positive sample set. For each fault issue, we used the models to select three descriptions or entry names that were most similar to the corresponding description or entry name but different. Finally, we constructed a sentence pair dataset with a ratio of 1:3 for positive to negative samples, intended for the Fault Description and FIM Manual Matching task.

\textbf{QA dataset, corpus set, curated corpus set, curated reorganized set}:We manually constructed 97 civil aviation fault question-and-answer pairs from failure cases, which cover aspects such as aircraft principles and fault reasoning. Combined with the 105 question-and-answer pairs from the exam, we formed a question-and-answer dataset. We gathered the textual content from textbooks, bilingual aligned corpora, FIM and TSM manuals, typical failure cases, and related articles, then performed semantic segmentation with chunk size set to 300, removing texts shorter than 20 characters. The segmented texts were combined with the answers from the question-and-answer dataset to form a corpus dataset. Given that the corpus contains expressions similar to or even better than the answers, we used a large model to assist in building a curated corpus set and a curated reordering set (see Appendix A for details).The four datasets are used for Civil aviation maintenance QA task.

\textbf{Multiple-choice dataset}:We constructed a large-scale multiple choice question dataset covering 12 aircraft types, comprising a total of 6,974 professional questions . Additionally, we curated 995 multiple choice questions sourced from open access online resources, covering seven specialized domains: aircraft structures and systems, aircraft airframe structure repair, aircraft component repair, aircraft maintenance and technical support technology, avionics systems, civil aircraft mechanic professional skills knowledge, and civil aircraft maintenance licensing examinations.The datasets is used for Civil aviation maintenance Multiple Choice task.

\textbf{Fault-tree dataset}:Although traditional evaluation methods, such as multiple choice accuracy tests, entry extraction, and fault system localization, are useful, they often fall short of assessing the deep reasoning capabilities of a LLM. To address this limitation, we have developed an innovative approach employing fault tree analysis. Our method involves constructing a structured evaluation set directly from authentic historical troubleshooting cases pertaining to B737 and A320 aircraft.
The inherent complexity of real-world maintenance scenarios, characterized by numerous implicit potential causes, presents significant challenges for models attempting multi-step reasoning. Furthermore, a single "correct" diagnostic path in a real-world scenario is often not exhaustive; alternative valid sequences of deductions can lead to the same fault phenomenon. This introduces considerable complexity when evaluating a model.
Our fault tree construction process meticulously analyzes each historical case to identify the root cause and all potential causes. These relationships are then systematically mapped into a hierarchical tree structure. Specifically, each case is transformed into a fault tree where the root node represents the observed fault phenomenon, intermediate nodes denote causes and sub-causes, and the ultimate root cause is situated at the leaf nodes. This structure provides a comprehensive representation of the real fault case.

\begin{table}
  \caption{Details of Datasets}
  \centering 
  \begin{tabular}{lllll}
    \toprule
    Task & Samples & 10th Percentile & Median & 90th Percentile\\ 
    \midrule
    Chinese-English aligned dataset & 1336 & 1 & 3 & 180 \\
    Categorize dataset & 100 & 2 & 15 & 41 \\ 
    Cluster dataset & 613 & 244.2 & 286 & 297 \\
    Sentence pairing dataset & 5984 & 6 & 20 & 234 \\
    QA dataset & 202 & 25 & 263 & 299 \\
    Multiple-choice dataset & 7969 & 3 & 8 & 18 \\
    Fault-tree dataset & 50 & 6 & 33 & 495 \\
    \bottomrule 
  \end{tabular}
  \label{tab:table}
\end{table}

\subsection{Task Construction}
We designed seven types of tasks, each used for evaluating Embeddings and LLMs, including:Alignment bilingual terminology,Aircraft fault system location,Aircraft text chapter location,Civil aviation maintenance Multiple Choice,Fault description and FIM manual Match,Civil aviation maintenance QA,Fault-tree QA.

\textbf{Alignment bilingual terminology}

For \textbf{Embeddings}, the task is BitextMining, which aims to measure the model's ability to map 'synonymous' sentences in different languages to similar vectors. We use a Chinese-English aligned dataset that includes text at various granularities such as words, sentences, and paragraphs. The dataset matches civil aviation English texts with their corresponding Chinese translations, and we evaluate the model's performance using F1 score.

For \textbf{LLMs},models were prompted to translate aviation-domain English sentences into Chinese while preserving original meaning and terminology, ensuring fidelity without introducing additional content. 
Translation quality is initially assessed using BLEU scores to measure n-gram overlap between generated and reference translations. We consider translations with BLEU scores greater than 0.05 as correct. 

\textbf{Aircraft fault system localization}

For \textbf{Embeddings and LLMs}, the task requires the model to correctly assign fault descriptions to corresponding aircraft systems, thereby evaluating the model's fault localization capability. We use accuracy as an indicator, calculated as the ratio of the number of correctly predicted samples to the total number of samples.

\textbf{Aircraft text chapter localization}:

For \textbf{Embeddings}, this task is Clustering, which measures the model's ability to cluster textual data of similar systems in civil aviation maintenance. We use the k-means algorithm (k=10) for text clustering and evaluate it using V-measure.

For \textbf{LLMs}, These include fault localization scenarios where the model must identify the appropriate primary system from a predefined candidate list based on textual fault descriptions.Evaluate using accuracy.

\textbf{Civil aviation maintenance Multiple Choice}:

\textbf{For Embeddings and LLMs}, the task aims to evaluate the model's ranking precision and discrimination ability. We assess using a multiple-choice dataset and measure accuracy.

\textbf{Fault description and FIM manual Match}:

\textbf{For Embeddings}, the goal is to measure the model's alignment capability on fine-grained corpora. We selected a sentence pair dataset. This is used to evaluate the model's ability to match failure questions with their corresponding failure descriptions and failure entries. Evaluation is done using F1 score.

\textbf{For LLMs}, these tasks require the model to link aircraft maintenance entries with their detailed explanations, as well as associate fault descriptions with the corresponding entries in the maintenance manuals.
Matching tasks are evaluated using standard accuracy, where models directly output binary predictions (0/1) indicating whether entities or concepts match correctly.

\textbf{Civil aviation maintenance QA}:

For \textbf{Embeddings}, the tasks is to evaluate the model's ability to retrieve relevant data. Each model's retrieval capability is assessed using a curated corpus and a curated query set. We convert the query set text and corpus text into vectors, retrieve the corpus based on the similarity of the query vector, and finally calculate ndcg@10 based on the retrieval results. Subsequently, we use the top 10 from the curated corpus for re-ranking tasks to test the model's fine-grained ranking ability, with the evaluation metric also being ndcg@10.

For \textbf{LLMs}, Models are required to answer technical questions about aircraft maintenance.Q\&A tasks are evaluated using LLM-as-a-judge with the same three-tier scoring system (0/1/2 points), where the evaluator determines the correctness of responses based on domain expertise.

\textbf{Fault Tree-structured QA}:

For \textbf{Embeddings}, since an answer from one sample might appear in the query of another sample, we gather all answers together and remove duplicates with the queries to obtain the corpus for each query. Then, the model retrieves texts based on the number of answers. If none of the retrieved texts contain any of the answers, it is marked as 0; otherwise, it is marked as 1. The final evaluation is conducted using accuracy.

For \textbf{LLMs}, a complete fault tree is segmented into a series of inference sub-tasks, each targeting a specific level within the reasoning hierarchy. In this task, the model is provided with the inference history from all preceding levels as contextual information. For subtasks at higher levels of difficulty, specific knowledge relevant to the current layer is also provided. This design allows the model to infer plausible nodes for the subsequent level based on the available historical information and knowledge, thereby enabling a robust assessment of its multi-step reasoning capabilities.We employ human evaluation with a three-tier scoring system. We compare LLM responses with the gold label using the following criteria: no match (0 point), partial match (1 point), and complete match (2 points).

\section{Experimental evaluation}
\vspace{-10pt}  
\subsection{Setting}
\vspace{-5pt}   

For Embeddings, we selected Conan-embedding-v1,gte-large-zh,m3e-large,BGE-large-zh-v1.5, gte-Qwen2-1.5B-instruct, gte-Qwen2-7B-instruct, Qwen3-Embedding-4B, and Qwen3-Embedding-8B for dataset evaluation.

For LLMs, we conducted comparative tests across six Qwen models (Qwen2.5-32B, Qwen3-32B, Qwen3-30B-A3B, QWQ-32B, Qwen3-235B-A22B and Qwen3-235B-Instruct) alongside three additional models(Claude-Opus-4, DeepSeek-R1-0528, and GPT-o3-mini). Two prompting strategies were employed: in thinking mode, models were instructed to analyze problems step by step before providing a final answer, while in non-thinking mode, models directly output the answer choice. For decoding, we follow official recommendations, for example, for all Qwen3 models in the thinking mode, we set a sampling temperature of 0.6, a top-p value of 0.95, top-k value of 20, and max output length to 32768 tokens. See Appendix A for more details of the setup.

\subsection{Results}
\subsubsection{Embedding}

\begin{table}[H]
  \caption{Results of Embedding}
  \centering 
  \small
  \begin{tabular}{lllcccccccc}
    \toprule
    Name & Size & \makecell{Mean\\(Task)} & \makecell{Bitext\\Mining} & \makecell{Classi-\\fication} & Cluster & \makecell{Pair\\Class} & \makecell{Retrieval} & \makecell{Reranker-\\choice} & \makecell{Reranker-\\text} & \makecell{Fault-\\ Tree} \\
    \midrule
    Conan-embedding-v1 & 326M & 55.14 & 49.46 & 29 & 76.55 & 69.13 & 55.06 & 28.51 & 78.28 & 18 \\
    gte-large-zh & 326M & 49.67 & 33.30 & 25 & 73.68 & 65.33 & 46.76 & 29.24 & 74.39 & 28\\ 
    m3e-large & 340M & 43.37 & 22.08 & 23 & 72.90 & 58.52 & 34.11 & 27.86 & 65.10 & 18\\
    BGE-large-zh-v1.5 & 671M & 55.14 & 44.82 & 31 & 78.65 & 61.11 & 63.27 & 30.19 & 77.58 & 22\\
    gte-Qwen2-1.5B-instruct & 1.5B & 59.60 & 68.07 & 18 & 79.67 & 68.07 & 71.18 & 30.59 & 82.29 & 26\\
    gte-Qwen2-7B-instruct & 7B & 63.42 & 80.25 & 24 & 83.73 & 68.28 & 71.26 & 32.38 & 84.75 & 30\\
    Qwen3-Embedding-4B & 4B & 62.95 & 81.37 & 22 & 78.26 & 69.17 & 71.52 & 29.76 & 88.85 & 30\\
    Qwen3-Embedding-8B & 8B & 66.27 & 80.79 & 33 & 87.34 & 69.69 & 67.62 & 30.46 & 95.32 & 32\\
    \bottomrule 
  \end{tabular}
  \label{tab:table}
\end{table}

\subsubsection{LLM}

\begin{table}[H]
 \caption{Reasoning Baseline Results}
  \centering
  \setlength{\tabcolsep}{5pt} 
  \small
  \begin{tabular}{@{}l *{7}{c}@{}}
    \toprule
    & GPT-o3-mini   & claude-opus-4     & \makecell{deepseek-\\R1-0528} &  \makecell{Qwen3-235B\\-A22B} & QwQ-32B & \makecell{Qwen3-\\30B-A3B} & \makecell{Qwen3-235B-A22B\\-Thinking-2507}\\
    \midrule
    737CL-ME & 47.21\% & 63.83\%  & 59.34\% & 66.61\%  & 63.11\%  & 59.34\% & 66.39\%\\
    737CL-AV & 53.86\% & 64.02\%  & 60.98\%  & 65.85\%  & 65.04\%  & 58.33\% & 66.26\%\\
    737NG-ME & 54.10\% & 64.64\%  & 61.66\%  & 69.44\%  & 66.56\%  & 61.55\% & 68.69\%\\
    737NG-AV & 54.51\% & 63.13\%  & 61.92\%  & 62.73\%  & 62.93\%  & 55.31\% & 66.13\%\\
    767ME & 57.19\% & 64.71\%  & 58.82\%  & 60.95\%  & 62.09\%  & 56.21\% & 61.27\%\\
    767AV & 50.38\% & 58.17\%  & 53.42\%  & 59.70\%  & 58.37\%  & 55.70\% & 58.94\%\\
    A320ME & 56.67\% & 69.47\%  & 65.27\% & 68.37\%  & 74.04\%  & 63.99\% & 72.39\%\\
    A320AV & * & 66.60\%  & 63.69\% & 67.38\%  & 70.10\%  & 63.50\% & 70.68\%\\
    D328 & 57.67\% & 62.78\%  & 57.19\% & 57.19\%  & 58.31\%  & 54.63\% & 59.11\%\\
    HAWKER800XP & * & 57.76\%  & 52.90\% & 50.28\% & 47.66\% & 48.60\% & 52.52\%\\
    PRIMIER & * & 63.06\%  & 56.58\% & 56.39\%  & 58.74\%  & 53.83\% & 57.17\%\\
    GIV & * & 66.61\%  & 58.26\% & 57.02\%  & 57.55\% & 55.06\% & 58.79\%\\
    Overall & * & 63.81\%  & 59.27\% & 62.21\%  & 62.24\% & 57.40\% & 63.44\%\\
    \bottomrule
  \end{tabular}
  \label{tab:reasoning_baseline}
\end{table}

\begin{table}[H]
 \caption{Non-thinking Baseline Results. The highest and second-best scores are shown in bold and underlined, respectively.}
  \centering
  \setlength{\tabcolsep}{3pt} 
  \small
  \begin{tabular}{@{}l *{7}{c}@{}}
    \toprule
    & Qwen2.5-32B  & \makecell{Qwen3-235B-\\A22B-Instruct}     & Kimi-K2 &  Qwen3-235B-A22B & Qwen3-32B & Qwen3-30B-A3B & \makecell{Qwen3-30B-\\A3B-Instruct}\\
    \midrule
    737CL-ME & 65.96\% & 75.29\% & 68.09\% & 70.87\% & 61.15\% &  61.15\% & 65.25\%\\
    737CL-AV & 61.18\% & 69.11\% & 64.43\% & 62.60\% & 59.55\% &  58.74\% & 61.38\%\\
    737NG-ME & 66.35\% & 76.25\% & 68.90\% & 70.82\% & 64.22\% &  62.51\% & 66.45\%\\
    737NG-AV & 57.72\% & 70.14\% & 68.74\% & 60.32\% & 58.32\% &  57.72\% & 63.73\%\\
    767ME & 60.95\% & 65.36\% & 64.87\% & 62.09\% & 60.46\% &  60.13\% & 63.56\%\\
    767AV & 57.98\% & 67.68\% & 60.84\% & 57.41\% & 53.99\% &  50.19\% & 55.70\%\\
    A320ME & 73.67\% &82.45\%   & 74.41\% &  71.85\%& 68.74\% &  62.71\% & 73.13\%\\
    A320AV & 73.40\% & 81.36\% & 76.70\% & 66.80\% &  69.32\%&  66.02\% & 71.26\%  \\
    D328 &  56.87\%& 62.94\% & 59.74\% & 53.99\% & 59.42\% &  51.44\% & 55.27\%\\
    HAWKER800XP & 51.96\% & 54.39\% & 56.26\% & 48.60\% & 50.09\% &  49.53\% & 51.03\%\\
    PRIMIER & 50.69\% & 57.17\% & 58.15\% & 55.40\% & 54.22\% & 52.06\% & 55.01\%\\
    GIV & 53.64\% & 59.50\% & 65.01\% & 57.02\% & 57.37\% &  52.58\% & 54.00\%\\
    Overall & 61.20\% & \textbf{68.87\%} & \underline{65.66\%} & 63.84\% & 60.81\% &  60.55\% & 62.27\%\\
    \bottomrule
  \end{tabular}
  \label{tab:non-thinking_baseline}
\end{table}

\begin{table}[H]
  \caption{Results of LLMs}
  \centering 
  \begin{tabular}{lllccccccc}
    \toprule
    Name & \makecell{Translation} & \makecell{System\\Localization} & \makecell{Chapter\\Localization} & \makecell{FIM Manual\\Match} & \makecell{maintenance\\QA} & \makecell{Reasoning\\on Tree} \\
    \midrule
    Claude-opus-4  & 69.01 & 71 & 92.01 & 82.07 & 44.80 & 36 \\ 
    deepseek-R1-0528   & 65.87 & 77 & 90.70 & 86.92 & 47.03 & 48 \\
    Qwen3-235B-A22B(T) & 67.00 & 68 & 90.05 & 79.41 & 52.72 & 38 \\
    QwQ-32B & 69.09 & 64 & 88.58 & 70.69 & 47.52 & 46 \\
    Qwen3-30B-A3B(T) & 67.59 & 68 & 86.95 & 78.21 & 47.52 & 30 \\
    Qwen2.5-32B-Instruct & 65.64 & 64 & 91.19 & 77.29 & 39.36 & 26 \\
    Qwen3-235B-A22B-Instruct  & 69.39 & 74 & 94.94 & 62.90 & 61.88 & 58 \\
    Kimi-K2 & 71.12 & 84 & 92.66 & 55.56 & 49.01 & 28 \\
    Qwen3-235B-A22B(N) & 69.61 & 69 & 87.44 & 69.20 & 57.67 & 44 \\
    Qwen3-32B(N) & 66.32 & 67 & 77.98  & 58.31  & 50.50 & 34 \\
    Qwen3-30B-A3B(N) & 66.92 & 63 & 78.30 & 41.64  & 46.53 & 34 \\
    Qwen3-30B-A3B-Instruct & 67.44 & 62 & 91.35 & 42.33 & 49.50 & 42 \\
    Qwen3-235B-A22B-Thinking-2507 & 68.71 & 78 & 92.82 & 49.28 & 53.47 & 50 \\
    \bottomrule 
  \end{tabular}
  \label{tab:[5_tasks]}
\end{table}

\section{Analysis}
\subsection{Embedding}
The evaluation of 8 embedding models on the dataset indicates that model performance has significant task dependency and scale relevance. Qwen3-Embedding-8B ranks first with an average task score of 66.27, which is 22.9 points higher than m3e-large (43.37). In terms of model size, the evaluation results show a threshold effect: when the model is smaller than 1 billion parameters, size is not the most critical factor affecting performance. However, when the model exceeds 1 billion parameters, there is a positive correlation between size and performance. On different tasks, BitextMining shows a dichotomy; Classification and Reranker-choice, as well as Fault-Tree models, perform poorly, possibly because these three tasks are more inclined towards knowledge embedding, while most embedding models primarily focus on semantic embedding.
To ensure the robustness of our evaluation protocol, we conducted a comprehensive replication study using tasks and datasets from MTEB that closely mirror the CAMB benchmark (see Appendix G for the full list). The resulting performance distributions and aggregate metrics were qualitatively consistent with those reported in the original CAMB study, corroborating the validity and generalizability of our evaluation framework.

\subsubsection{Abnormal effects in retrieval tasks}

\begin{figure}[h]
  \centering
  \includegraphics[width=0.8\textwidth]{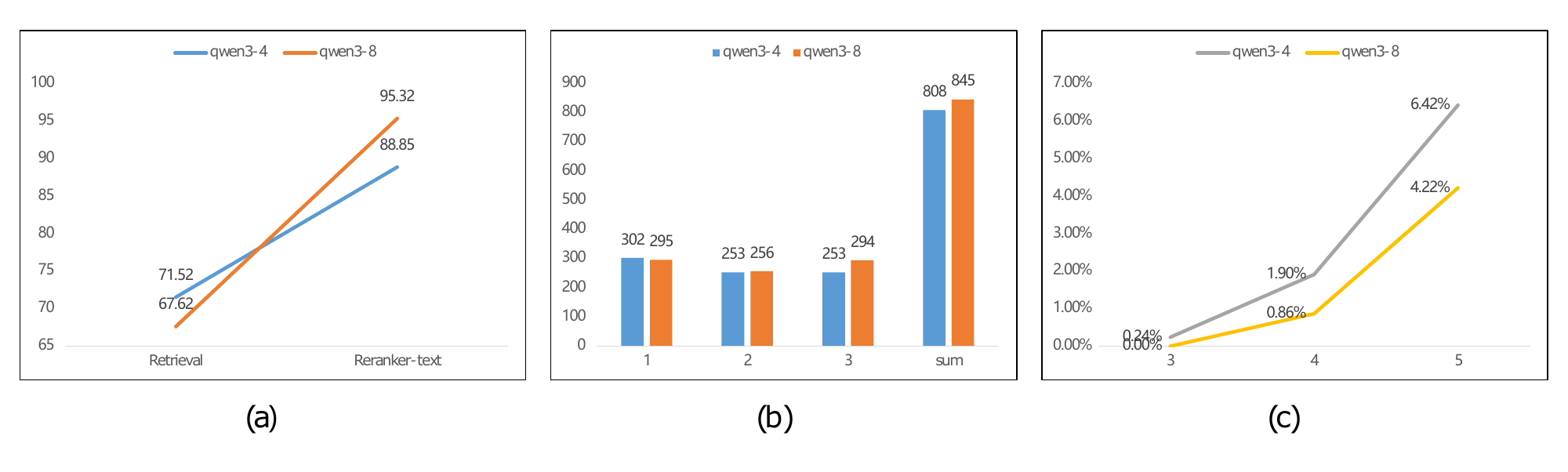}
  \caption{Retrieval tasks analysis.}
  \label{fig:fig3}
\end{figure}

Figure (a) shows that in Retrieval, Qwen3-8B is 4\% lower than Qwen3-4B, but in Reranker-Text, Qwen3-8B is 8\% higher than Qwen3-4B, indicating that Qwen3-8B has stronger ranking capabilities but may retrieve more diverse content. Figure (b) displays the number of low-relevance texts retrieved by both models, revealing that Qwen-8B retrieves more low-relevance content, which might be the reason for its poorer performance in Retrieval. Figure (c) illustrates the situation where low-relevant content is ranked within the top 3-5 by the models among their own retrieved content, suggesting that Qwen3-4B tends to place low-relevant content at the forefront, which also explains why Qwen-4B performs worse than Qwen3-8B in Reranker-Text.From this, it can be seen that Qwen3-8B is more aggressive than Qwen3-4B.

\subsubsection{Further analysis of the Fault tree}
\begin{figure}[h]
  \centering
  \includegraphics[width=0.7\textwidth]{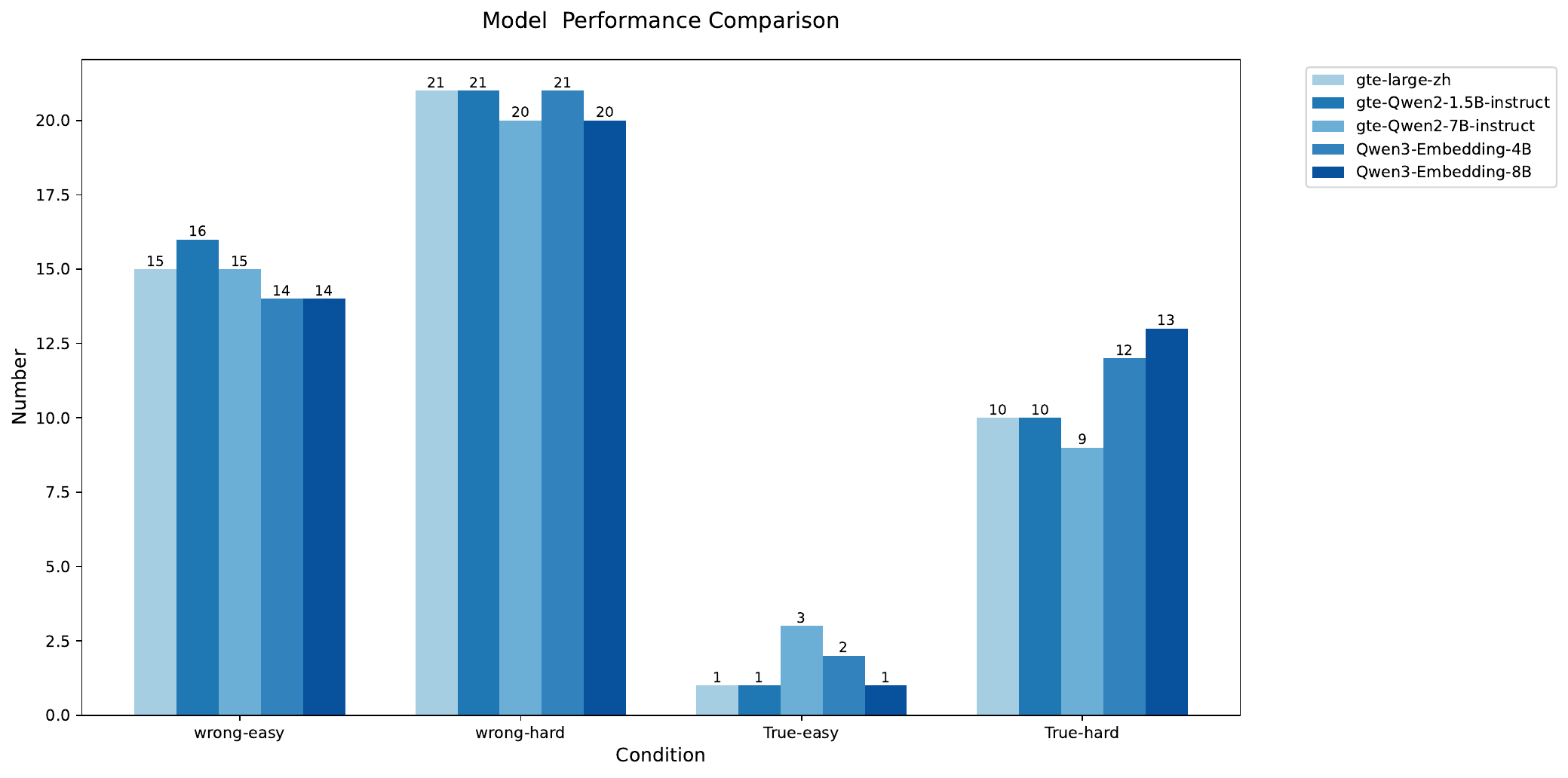}
  \caption{analysis of the Fault tree}
  \label{fig:fig4}
\end{figure}

Given that the fault tree dataset only has 50 entries, to more robustly evaluate the model's results, we assess the models from three dimensions, prioritized from (1) to (3).For the definition of question difficulty, see the appendix:

(1) The fewer completely wrong questions, the better;

(2) Among the completely wrong questions, the fewer simple ones, the better;

(3) Among the not completely wrong questions, the more difficult ones, the better.

From the first dimension, gte-large-zh, gte-Qwen-2-1.5B-instruct, gte-Qwen-2-7B-instruct, Qwen3-Embedding-4B, and Qwen3-Embedding-8B performed best, correctly answering 14, 13, 15, 15, 16 questions respectively. The differences between the five models are less than 3, indicating a consistent level, and thus the second dimension evaluation was conducted. Figure 4 shows that in the second dimension, Qwen3-Embedding-8B = Qwen3-Embedding-4B > gte-Qwen-2-7B-instruct > gte-large-zh > gte-Qwen-2-1.5B-instruct. In the third dimension, Qwen3-Embedding-8B > Qwen3-Embedding-4B > gte-Qwen-2-1.5B-instruct = gte-large-zh > gte-Qwen-2-7B-instruct. Overall, Qwen3-Embedding-8B > Qwen3-Embedding-4B > gte-Qwen-2-7B-instruct > gte-large-zh > gte-Qwen-2-1.5B-instruct.

\subsubsection{Models Efficiency}
\begin{figure}[h!]
  \centering
  \includegraphics[width=0.6\textwidth]{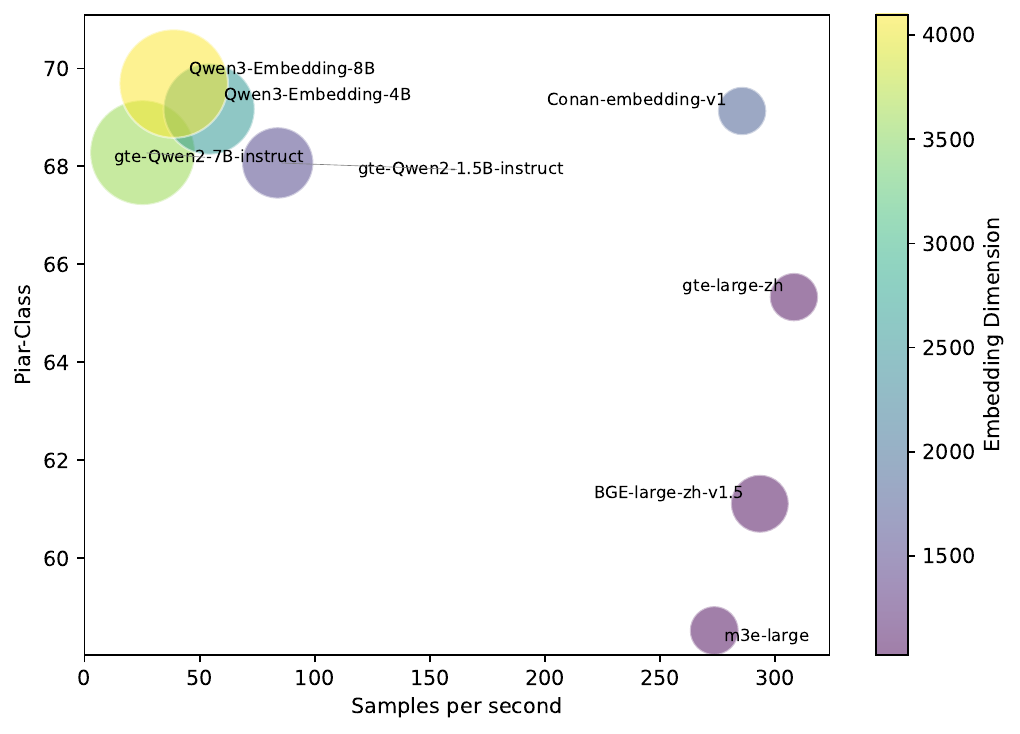}
  \caption{Model-efficiency}
  \label{fig:fig5}
\end{figure}
In the benchmark tests, the evaluated models differ in terms of parameter scale, model size, speed, and performance. The quality of a model depends on specific needs, so we visualized the speed, model size, embedding size, and score of each model in Figure 5. We chose the sentence pair task due to its large dataset and more noticeable model comparisons. From the graph, it can be seen that each model has its own strengths and weaknesses. Larger models generally have larger embeddings but slower speeds and better performance. It can be observed that Conan-embedding-v1 maintains a balance between performance and speed.

\subsection{LLM}

From Tables \ref{tab:reasoning_baseline} and \ref{tab:non-thinking_baseline}, we observe several key insights from the performance comparison between multiple choice aircraft maintenance datasets: (1) Qwen3-235B-Instruct delivers the highest overall precision (68. 87\%), significantly outperforming other models with a 3.2 percentage point advantage over the next-best model (Kimi-k2 at 65.67\%).(2)Regarding reasoning modes, the impact varies. For the smaller Qwen3-32B, thinking mode provides 1.21 percentage point gain versus no-thinking mode. However, this advantage diminishes in the larger Qwen3-235B, where thinking mode (62.22\%) shows only marginal improvement over non-thinking mode (62.04\%). (3)Performance patterns across aircraft types reveal critical domain-specific challenges. All models excel on A320 systems, particularly in mechanical (ME) and avionics (AV) subsystems where Qwen3-235B-Instruct achieves exceptional 82.45\% and 81.36\% accuracy respectively. On the contrary, HAWKER800XP presents consistent difficulties, with no model exceeding 60\% accuracy. 

Table \ref{tab:[5_tasks]} compares 13 language models across six evaluation metrics: Translation, System Localization, Chapter Localization, FIM Manual Match, Maintenance QA, and Reasoning on Tree. Key observations:

\textbf{Top Performers by Category:}
Analysis of the results shows different models excelling in specific domains. Kimi-K2 demonstrates superior performance in Translation (71.12) and System Localization (84), while Qwen3-235B-A22B-Instruct leads in three categories: Chapter Localization(94.94), Maintenance QA (61.88), and Reasoning on Tree (58). 

\textbf{Consistent High Performers:}
The Qwen3-235B family of models, particularly the Instruct version, emerges as the most capable overall performer. This model family shows particular strength in complex tasks requiring instruction following and reasoning capabilities.

\textbf{Parameter Size and Model Variant Effects:}
The results suggest clear benefits from larger model sizes, with 235B parameter models generally outperforming their 30B/32B counterparts. Instruct-tuned versions consistently outperform their base models, particularly in complex tasks. The "Thinking" variant shows mixed results, performing well in some metrics but lagging in others compared to similar-sized models.

These findings propose a deeper question: Why does the 235B model exhibit a less pronounced improvement under the thinking strategy than expected? To explore this phenomenon in greater depth, we conducted additional experiments across nine subjects within the CMMLU benchmark. Utilizing the decoding parameters recommended by the official implementation, our results generally demonstrate a consistent upward trend in performance when applying the thinking strategy and increasing the context length. Aside from the constraint that 1024 tokens are insufficient for our tasks, the majority of data points with thinking mode outperform those without it.

\begin{figure}[H]
  \centering
  \includegraphics[width=0.7\textwidth]{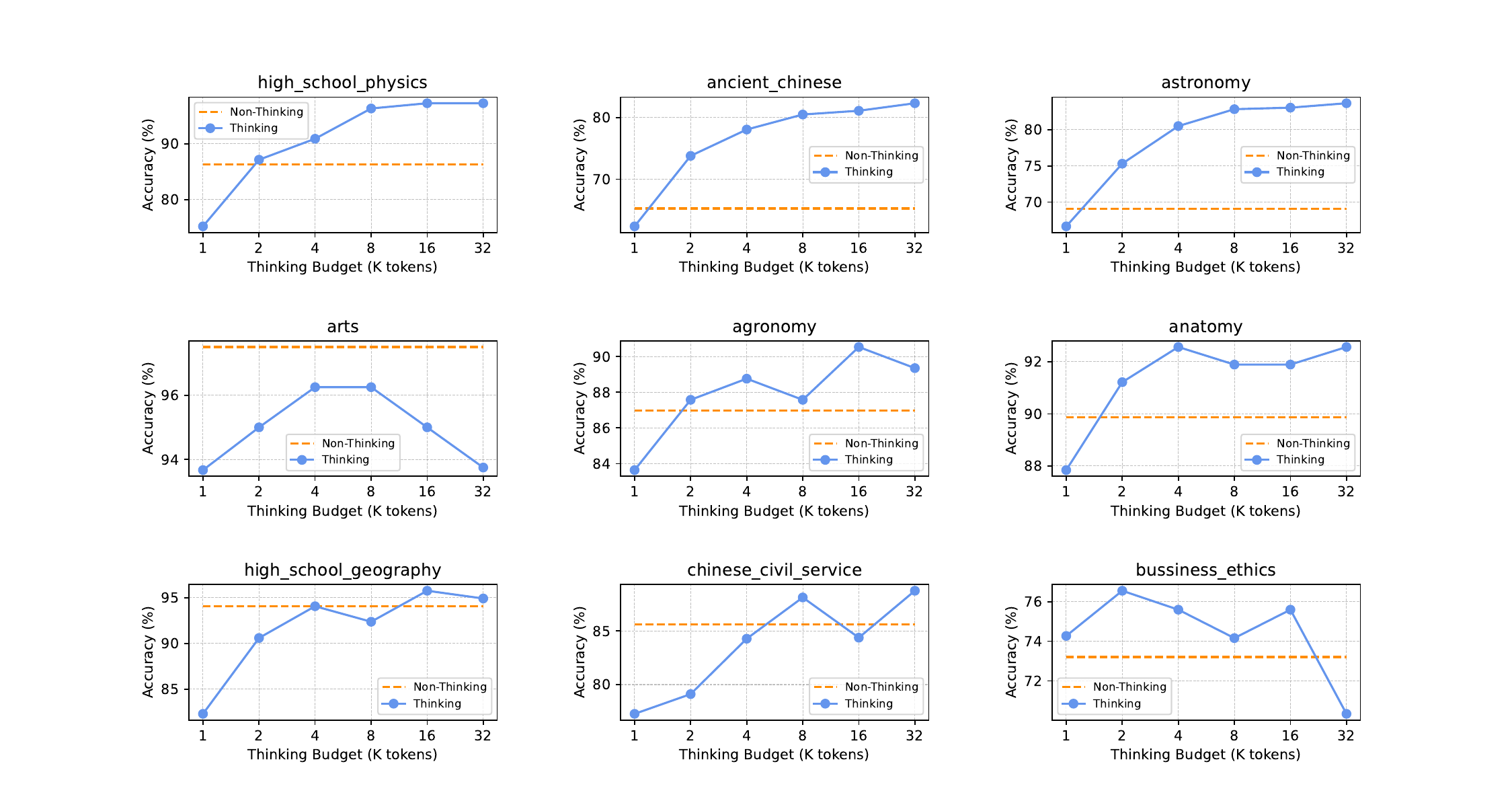}
  \caption{Performance Trends on 9 CMMLU Datasets}
  \label{fig:fig5}
\end{figure}

In our multiple choice questions, the thinking mode exhibits observable Test-Time Scaling Law properties; nevertheless, we conjecture that knowledge deficiencies and conceptual ambiguities likely cause it to underperform the instant responses of the non-thinking mode.

\begin{figure}[H]
  \centering
  \includegraphics[width=0.7\textwidth]{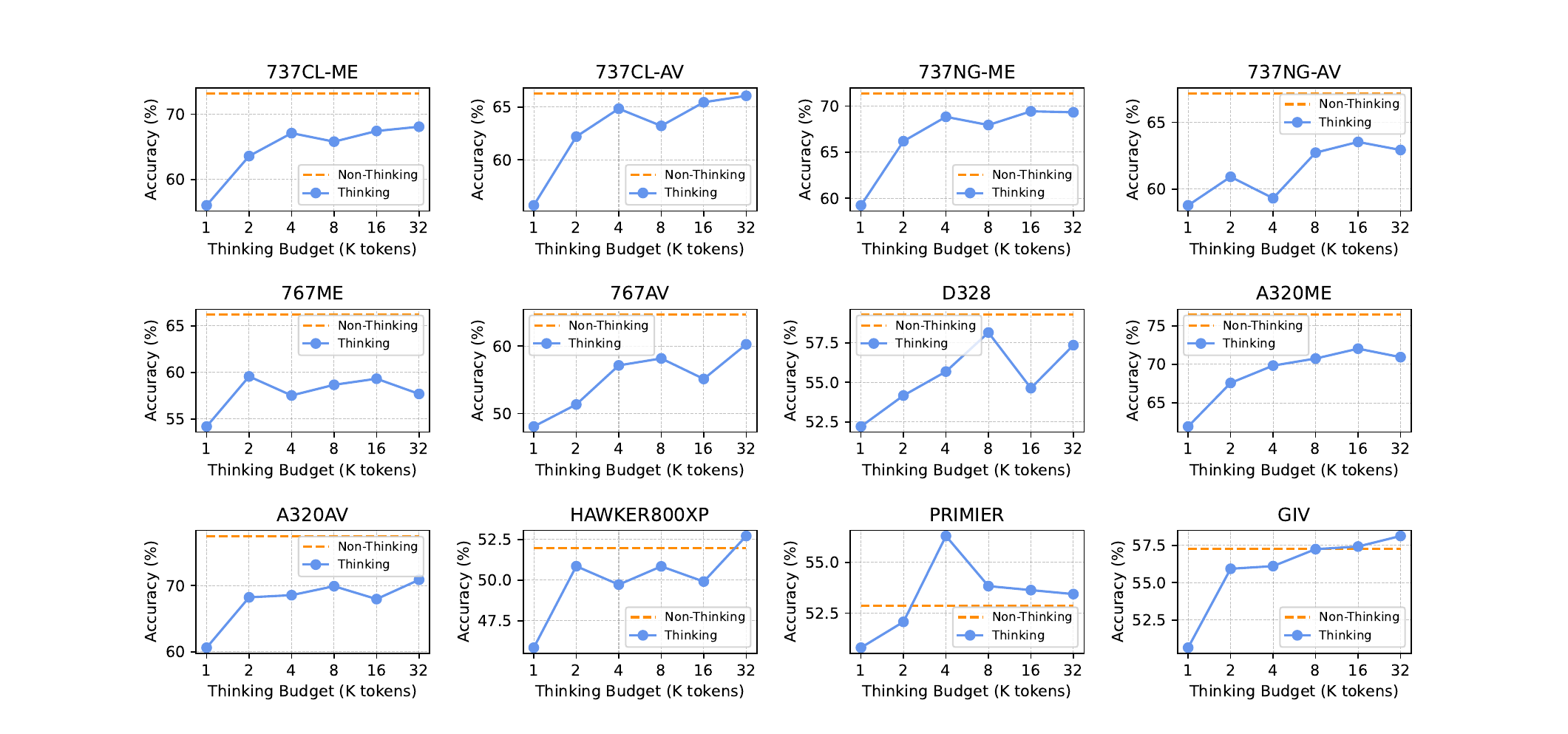}
  \caption{Performance Trends on 12 Aircraft Models}
  \label{fig:fig5}
\end{figure}
\section{Conclusion}
In conclusion, our research makes contributions to the LLMs and embedding models in the field of civil aviation maintenance. Firstly, we have proposed and constructed a dedicated LLM benchmark for civil aviation maintenance, with specific evaluation tasks detailed in Table 1, which fills the gap in specialized evaluation tools for this domain.
Through the evaluation of state-of-the-art embedding models and LLMs using this benchmark, several key insights have emerged. Current embedding models, while strong in semantic similarity, show limited accuracy in retrieving factual knowledge specific to civil aviation maintenance. For LLMs, their performance on multiple-choice tasks within the benchmark ranges from 60\% to 70\%. In reasoning-intensive tasks, despite exhibiting certain Test-Time Scaling Law properties due to knowledge deficiencies and conceptual ambiguities, they fail to significantly outperform the instant answering capability of non-thinking modes. Finally, these results validate that our benchmark is effective in distinguishing the performance of different embedding models and LLMs, confirming its utility for future research.

\section{Future Work}
Both embedding models and large language models (LLMs) show significant potential yet face distinct limitations in civil aviation maintenance. Our future work targets two complementary areas: enhancing knowledge retrieval capabilities and developing deep reasoning methods. Embedding models still struggle on multi-hop retrieval tasks. LLMs, while exhibiting a scaling law trend within the domain, have not yet surpassed the accuracy of non-thinking models. We hypothesize these deficiencies originate from inadequate domain-specific knowledge integration, and a fundamental mismatch between the deep reasoning patterns intrinsic to civil aviation maintenance and those prevalent in general-purpose reasoning. Overcoming this knowledge deficit and specifically adapting models to maintenance-centric deep reasoning constitute the primary objectives of our future work.
\clearpage

\appendix
\section*{Appendix}
\section{List of Evaluation Models and Sampling Parameters}
\begin{table}[h]
\centering
\caption{The Model List Used for Evaluation}
\label{tab:model_config}
\begin{tabular}{ll}
\toprule
Model & Published Date \\
\midrule
OpenAI-o3-mini & January 31, 2025 \\
Claude-opus-4 & May 22, 2025 \\
Deepseek-R1-0528 & May 28, 2025 \\
Qwen3-235B-A22B & April, 2025 \\
QwQ-32B & March 6, 2025  \\
Qwen2.5-32B-Instruct & September 19, 2024  \\
Qwen3-235B-A22B-Instruct & July 22, 2025  \\
Kimi-K2 & July, 2025 \\
Qwen3-32B & April, 2025  \\
Qwen3-30B-A3B & April, 2025  \\
Qwen3-30B-A3B-Instruct & July 30, 2025  \\
Qwen3-235B-A22B-Thinking-2507 & July, 2025 \\
\bottomrule
\end{tabular}
\end{table}
\begin{enumerate}
    \item \textbf{OpenAI-o3-mini}:We configured the parameter \texttt{max\_completion\_tokens} to 32768 while maintaining all other parameters at their default values.
    \item \textbf{Claude-opus-4}:We configured the model with max\_tokens set to 32,768 and \texttt{top\_p} at 0.95. Additionally, we enabled the thinking mode with a token budget of 16,384.
    \item \textbf{Deepseek-R1-0528}:We employed the following parameter configuration: \texttt{temperature} = 0.6, \texttt{max\_tokens} = 32768, \texttt{top\_p} = 0.95, \texttt{top\_k} = 20, while maintaining default values for all unspecified parameters.
    \item  For all \textbf{Qwen3} models in thinking mode, we set a sampling temperature of 0.6, a \texttt{top-p} value of 0.95, \texttt{top-k} value of 20, and max output length to
32768 tokens. 
    \item For \textbf{Kimi-K2}, all \textbf{Qwen2.5} and \textbf{Qwen3} models in non-thinking mode, the following parameter configuration was applied: \texttt{temperature} = 0.7, \texttt{max\_tokens} = 32768, \texttt{top\_p} = 0.8 and \texttt{top\_k} = 20 to balance token selection, along with \texttt{presence\_penalty} = 1.5, while keeping all unspecified parameters at their default settings. 
\end{enumerate}

\begin{figure}[H]
  \centering
  \includegraphics[width=0.7\textwidth]{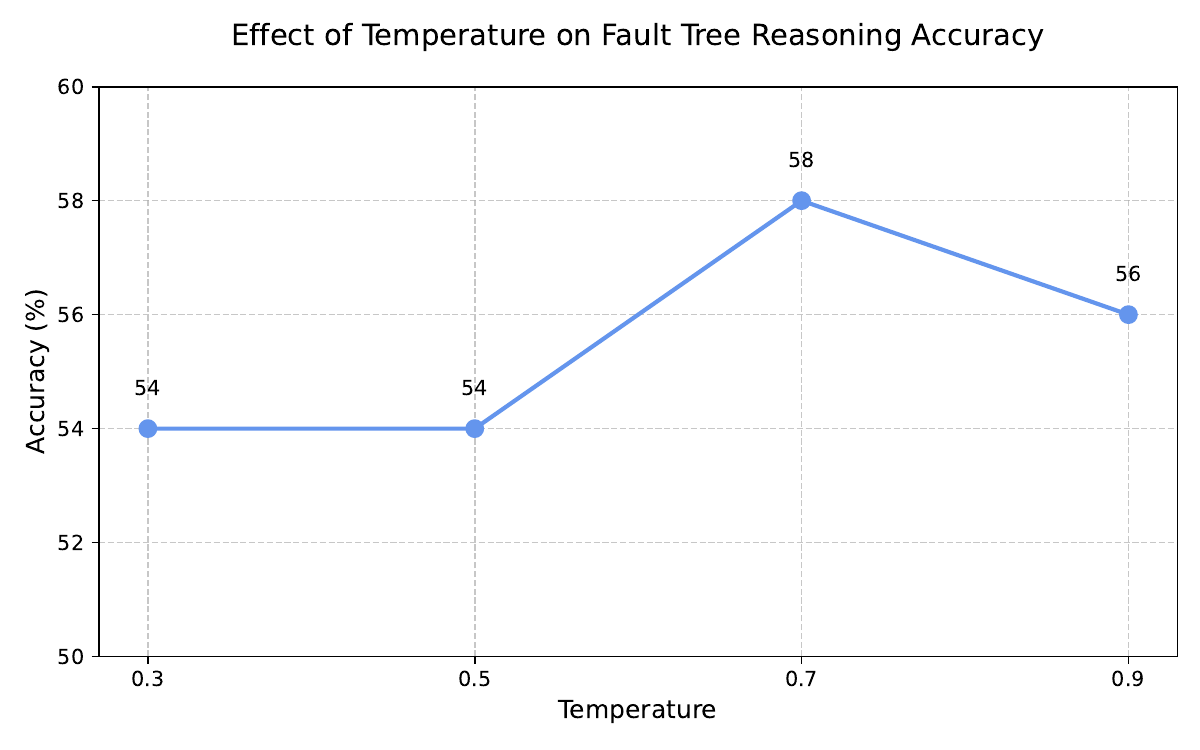}
  \caption{Model-Temperature Sensitivity Curve on Fault Tree}
\end{figure}

\section{Prompts}
    For translation, we use the system prompt "You are a translation specialist in the field of civil aviation." For other tasks, we use the system prompt "You are a specialist in the field of civil aircraft maintenance."
\subsection{Prompts for Multiple-Choice Question Task}
We designed two distinct prompt templates for \texttt{thinking} and \texttt{non-thinking} modes:
\subsubsection{Multiple-Choice Task}
\begin{enumerate}[leftmargin=*,label=\textbf{Mode \arabic*}:]
    \item \textbf{Non-thinking mode}
    \begin{mdframed}[backgroundcolor=promptbg,skipabove=3pt,skipbelow=3pt]
    Below is a multiple-choice question about civil aviation maintenance: 
    
    \textbf{Question:} Where is the DC power control component (1VE) installed?\\
    \hspace*{5mm}A. Under cabin floor \\
    \hspace*{5mm}B. Left electronic equipment bay \\
    \hspace*{5mm}C. Under floor at boarding door entrance \\
    \hspace*{5mm}D. Aft electronic equipment bay
    
    Answer directly and output only the correct option letter without any additional content.
    \end{mdframed}

    \item \textbf{Thinking mode}
    \begin{mdframed}[backgroundcolor=promptbg,skipabove=3pt,skipbelow=3pt]
    Below is a multiple-choice question about civil aviation maintenance: 
    
    \textbf{Question:} Where is the DC power control component (1VE) installed?\\
    \hspace*{5mm}A. Under cabin floor \\
    \hspace*{5mm}B. Left electronic equipment bay \\
    \hspace*{5mm}C. Under floor at boarding door entrance \\
    \hspace*{5mm}D. Aft electronic equipment bay
    
    Carefully consider and reason through your response, but ultimately output only the correct option letter.
    \end{mdframed}
\end{enumerate}
For the sake of brevity, only the non-thinking mode prompt is listed for subsequent tasks. The thinking mode prompt is generated by simply adding an instruction for careful consideration to the non-thinking prompt.
\subsubsection{Translation}

    \begin{mdframed}[backgroundcolor=promptbg,skipabove=3pt,skipbelow=3pt]
    Translate the following English to Chinese exactly, outputting only the translation without additions.\\
    Engilsh: keel beam
    \end{mdframed}
\subsubsection{System Localization}
    \begin{mdframed}[backgroundcolor=promptbg,skipabove=3pt,skipbelow=3pt]
    Based on the description, locate the faulty primary system. You must select exactly 1 option from the given candidate primary system list. Do not provide any other content.\\
    Description: The transit cabin crew reported a kerosene odor in the cabin during the descent phases.\\
    Candidate Primary System List: air conditioning, fuel...
    \end{mdframed}
\subsubsection{Fault description and FIM manual Match}
    \begin{mdframed}[backgroundcolor=promptbg,skipabove=3pt,skipbelow=3pt]
     
     DPlease determine whether the content following the text matches the FIM Entry/Fault Description. Respond with 1 if it matches, 0 if not. Output ONLY 0 or 1.\\
     FIM Entry/Fault Description: autothrottle arm problem \\
     Description: these observed faults: (a) the autothrottle does not arm. (b) arm does not show on the fma. (c) the arm light on the mcp does not come on. Note: the arm light on the mcp operates differently on the honeywell and rockwell collins mcps. The arm light on the honeywell mcp is not as bright as the rockwell collins mcp in daylight/sunlight conditions. the arm on the honeywell mcp does not come on when the master test and dimming switch is set to the test position.;(2) this task is for these maintenance messages: (a) 22-31541 a/t engage hold on 1) the flight control computer a (fcc-a) detects a discrete wraparound failure for the auto throttle (a/t) engage hold on or a software requested disengage did not occur during the bite. 2) associated interface pins: (j1a-j12) - output;(3) an outside air temperature (oat) value in the takeoff reference page is necessary for the a/t to arm. (a) airplanes with aspirated total air temperature (tat) probe: 1) entry of the oat is automatic. (b) airplanes with non-aspirated tat probe: 1) manual entry of the oat is necessary.
    \end{mdframed}

\subsubsection{Fault Tree}
    \begin{mdframed}[backgroundcolor=promptbg,skipabove=3pt,skipbelow=3pt]
    Based on the historical troubleshooting path including fault descriptions and other information: On April 5, the crew reported inaccurate radar echoes with displays showing no actual weather, and large cumulus clouds still appearing even with gain at minimum. The control panel and transceiver were replaced. During a follow-up flight, the Navigation Display showed full weather throughout under clear conditions. At the next stopover, charring was found at the drive unit and antenna interface, leading to replacement of the antenna and drive unit. On April 9, the crew reported "WX RADAR WORKING POOR". Knowledge: The weather radar system consists of radar control panel, transceiver, antenna, and drive unit working together for scanning and display. Ablation in antenna/drive unit may cause abnormal signal transmission affecting echo display (AMM 34-45-00, FIM 34-45-00). Provide the next most probable causes, not exceeding five.
    \end{mdframed}
\subsubsection{Prompts for gpt-4o to Evaluate Fault Tree Results}
        \begin{mdframed}[backgroundcolor=promptbg,skipabove=3pt,skipbelow=3pt]
    Please perform the following two tasks based on the question:

Content Coverage Judgment: Evaluate whether 'Answer 1' basically covers the core content in the 'Answer 2' list.

If yes, output 1

Otherwise, output 0

Specific Coverage Point Identification: Identify the specific entry numbers in 'Answer 2' that are covered by 'Answer 1'.

If none, output none.\\
\#\#\# Answers to Evaluate \#\#\# \\
Answer 1: {extracted}\\
Answer 2: (Reference Answer List): \\
\#\#\# Output Format Requirements \#\#\#\\
The output must strictly adhere to the following format:\\
First line: '1 or 0'\\
Second line: 'The specific entry numbers covered in Answer 2 (comma-separated, e.g. '2,3,5') or `none`\\
Note: Do not include any additional explanations or spaces in the output - only output two lines of text."\\

    \end{mdframed}
\subsubsection{Prompts for gpt-4o to Evaluate QA Results}
    \begin{mdframed}[backgroundcolor=promptbg,skipabove=3pt,skipbelow=3pt]
    Please perform the following task based on the question:
Check the relationship between Answer 1 and Answer 2:\\
If Answer 1 essentially covers the core content of Answer 2, output 2\\
If Answer 1 contains partial information from Answer 2, output 1\\
If Answer 1 contains no information from Answer 2, output 0\\
Answer 1: \{model's answer\}\\
Answer 2: \{ground\_truth\}\\
Note: Do not include any additional explanations or spaces in the output; only output the number.
    \end{mdframed}
\subsection{Prompts for paragraph sorting}
    \begin{mdframed}[backgroundcolor=promptbg,skipabove=3pt,skipbelow=3pt]
    Task: Sort paragraphs based on their relevance to the question, with the most relevant paragraph at the top of the output list. 
    
    Output: Return a list of paragraph indices (the first index in the list represents the most relevant paragraph, and the last index represents the least relevant paragraph). Note: Only output the list of indices! The paragraph indices start from 1.
    
    Example  : Question : Key points for engine maintenance?    Paragraphs : [Paragraph A, Paragraph B, Paragraph C]
    Example output : [1,3,2](indicates that paragraph 1 is the most relevant, paragraph 3 is the second most relevant,and paragraph 2 is the least relevant.)
    \end{mdframed}
 
\section{The Example of Construction of Fault Trees}
We first used LLMs to extract the fault tree structure from a fault case, and then humans corrected each tree.
\subsection{Original Fault Case}
\begin{mdframed}[backgroundcolor=promptbg,skipabove=3pt,skipbelow=3pt]
April 5\\
Crew reported inaccurate radar returns – weather displayed despite clear conditions. Full cumulus shown even at minimum gain. Replaced control panel and transceiver. Subsequent flight crew reported persistent full weather display on ND during clear weather. Next stopover revealed burn marks at drive unit/antenna interface. Replaced antenna and drive unit.

April 9\\
Crew reported "WX RADAR WORKING POOR." Reinstalled transceiver and cleaned connectors. Flight observer later reported actual scanning angle deviated downward by 3 degrees. Post-flight inspection showed normal antenna angle.

April 14\\
Recurring inflight weather radar fault with WXR FAIL code. Post-flight checks: wiring measured and transceiver swapped to B2635 – fault transferred.

Analysis:\\
April 5: Gain knob "inoperability" attributed to automatic gain adjustment in MAP mode. Control panel and transceiver functional. Antenna/drive unit burn marks caused inaccurate returns. No recurrence post-replacement.

April 9: Scanning angle discrepancy confirmed. Two possibilities: crew/observer error, or genuine drive unit defect (noted in other fleets). New drive unit initially suspected; no further reports warranted observation. (Contingency: recurring issues would require antenna angle tests and part replacement).

April 14: Distinct WXR FAIL fault. Isolated to transceiver (S/N 1FDYR) post-transfer to B2635. Resolution confirmed with no subsequent faults on B5026.
\end{mdframed}

\subsection{From Raw Extraction to Refined Tree Structures}
\begin{mdframed}[backgroundcolor=promptbg,skipabove=3pt,skipbelow=3pt]
\begin{verbatim}
{
      "fault_tree": {
        "root": {
          "id": "1-1",
          "description": "On April 5, the crew reported inaccurate radar echoes,
          with displays showing weather when there was none, 
          and large cumulus clouds still displayed even at minimum gain. 
          The panel and transceiver were replaced. 
          Subsequent flights reported full weather displays 
          on ND under clear conditions.
          During the next stop, blackening was 
          found at the drive unit and antenna interface, 
          leading to replacement of the antenna and drive unit. On April 9, 
          the crew reported 'W/X RADAR WORKING POOR'.",
          "knowledge": [
            "The weather radar system consists of the radar control panel, 
            transceiver, antenna, and drive unit, 
            which work together to achieve radar scanning and image display.",
            "Burn marks on the antenna and drive unit may cause abnormal 
            radar signal transmission, affecting echo display.",
            "AMM 34-45-00, FIM 34-45-00"
          ],
          "children": [
            {
              "id": "2-1",
              "description": "Reinstalled the transceiver and cleaned the connectors. 
              Subsequent flights reported the actual scan angle was 3 degrees lower. 
              On April 14, 'WXR FAIL' was reported again, with a WXR FAIL code.",
              "children": [
                {
                  "id": "3-1",
                  "description": "Post-flight line testing and swapping the transceiver 
                  to B26XX resulted in the fault transferring.",
                  "evidence": [
                    {
                      "navigate": "The fault transferred to B26XX, 
                      indicating an issue with the original transceiver.",
                      "next": "4-1"
                    }
                  ],
                  "children": [
                    {
                      "id": "4-1",
                      "description": "Replaced transceiver with serial number 1FDYR.",
                      "reason": "Fault code WXR FAIL and the fault transferring with 
                      the transceiver indicate a transceiver issue.",
                      "evidence": [
                        {
                          "result": "B50XX has not reported any further radar issues.",
                          "is_cause": true
                        }
                      ]
                    }
                  ]
                },
                {
                  "id": "3-2",
                  "description": "The actual scan angle being 3 degrees lower may 
                  indicate a drive unit defect or crew operation issue.",
                  "evidence": [
                    { "result": "The actual scan angle was problematic.", 
                    "is_cause": true }
                  ],
                  "children": [
                    {
                      "id": "4-2",
                      "description": "The drive unit was new; initial occurrence 
                      required further observation.",
                      "evidence": [
                        {
                          "result": "No further reports in the following days.",
                          "is_cause": false
                        }
                      ]
                    },
                    {
                      "id": "4-3",
                      "description": "If the issue persists, conduct antenna angle 
                      testing and consider replacement.",
                      "evidence": [
                        {
                          "result": "After replacement, the fault did not recur.",
                          "is_conclusion": true
                        }
                      ]
                    }
                  ]
                }
              ]
            }
          ]
        }
      }
    }
\end{verbatim}
\end{mdframed}

\section{Consistency Metrics (Human vs. GPT-4o)}
\begin{figure}[htbp]
    \centering
    \begin{subfigure}[b]{0.48\textwidth}
        \includegraphics[width=\linewidth]{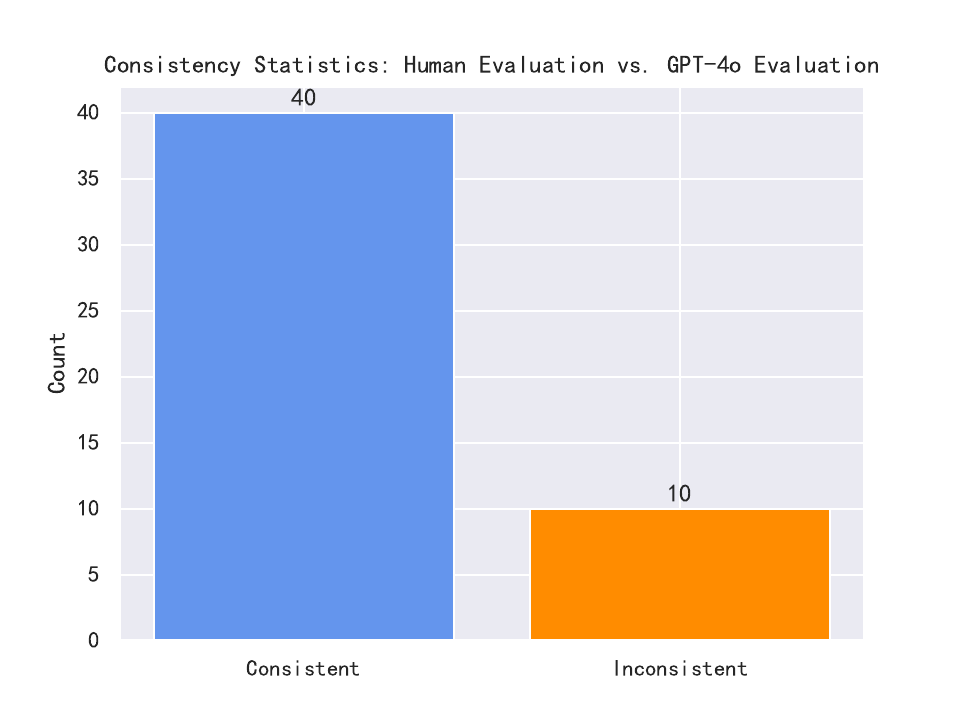} 
        \caption{Qwen2.5-32B-Instruct Consistency Metrics}
        \label{fig:qwen2.5-32b-consistency.pdf}
    \end{subfigure}
    \hfill 
    \begin{subfigure}[b]{0.48\textwidth}
        \includegraphics[width=\linewidth]{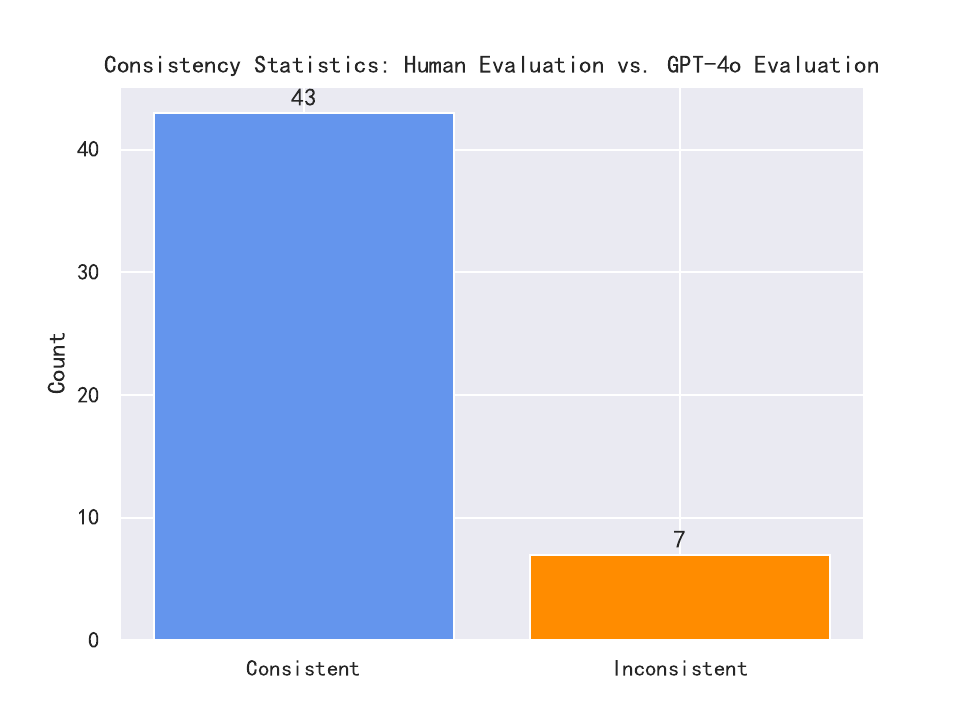} 
        \caption{Qwen3-235B-A22B-Instruct Consistency Metrics}
        \label{fig:consis}
    \end{subfigure}
    \caption{Human vs. GPT-4o Consistency in Fault Trees: LLM-0/LLM-1 Discrepancy Counts}
    \label{fig:parallel}
\end{figure}

\begin{figure}[htbp]
    \centering
    \begin{subfigure}[b]{0.48\textwidth}
        \includegraphics[width=\linewidth]{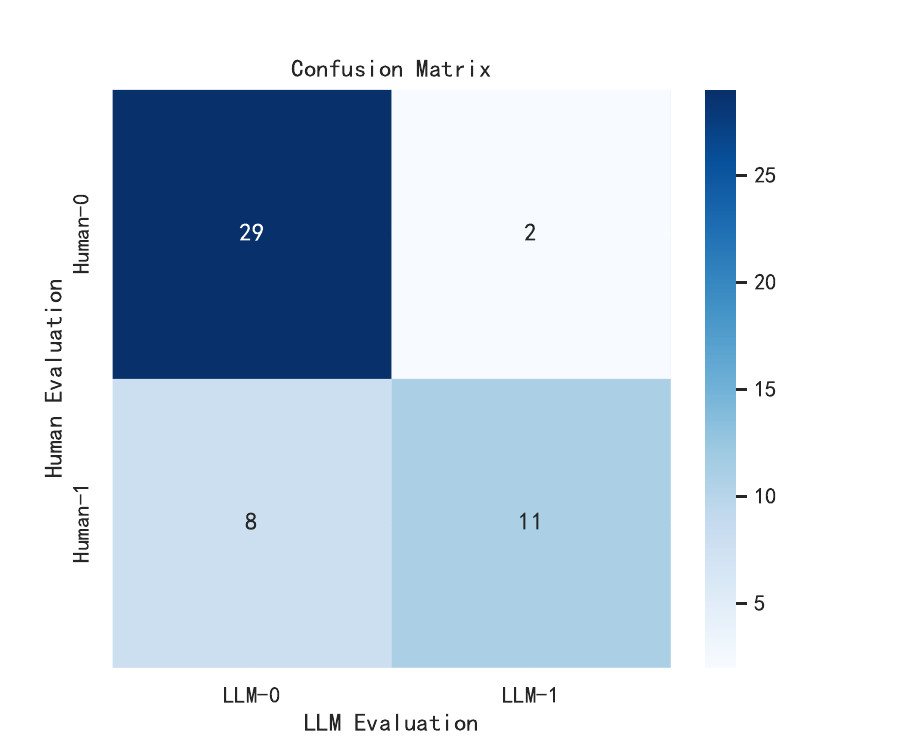} 
        \caption{Qwen2.5-32B-Instruct Consistency Metrics}
        \label{fig:qwen2.5-32b-consistency.pdf}
    \end{subfigure}
    \hfill
    \begin{subfigure}[b]{0.48\textwidth}
        \includegraphics[width=\linewidth]{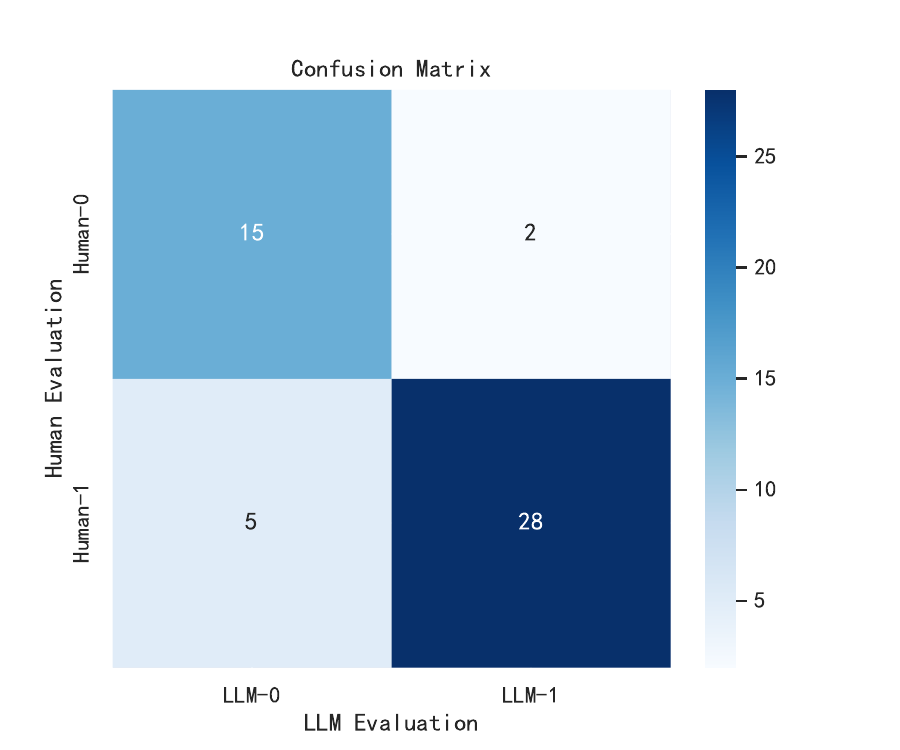} 
        \caption{Qwen3-235B-A22B-Instruct Consistency Metrics}
        \label{fig:consis}
    \end{subfigure}
    \caption{Consistency Metrics: Human vs. GPT-4o Evaluation of Fault Trees via Confusion Matrices}
    \label{fig:parallel}
\end{figure}
\clearpage

\section{Curated Corpus Set Construction}
\begin{figure}[h!]
  \centering
  \includegraphics[width=1\textwidth]{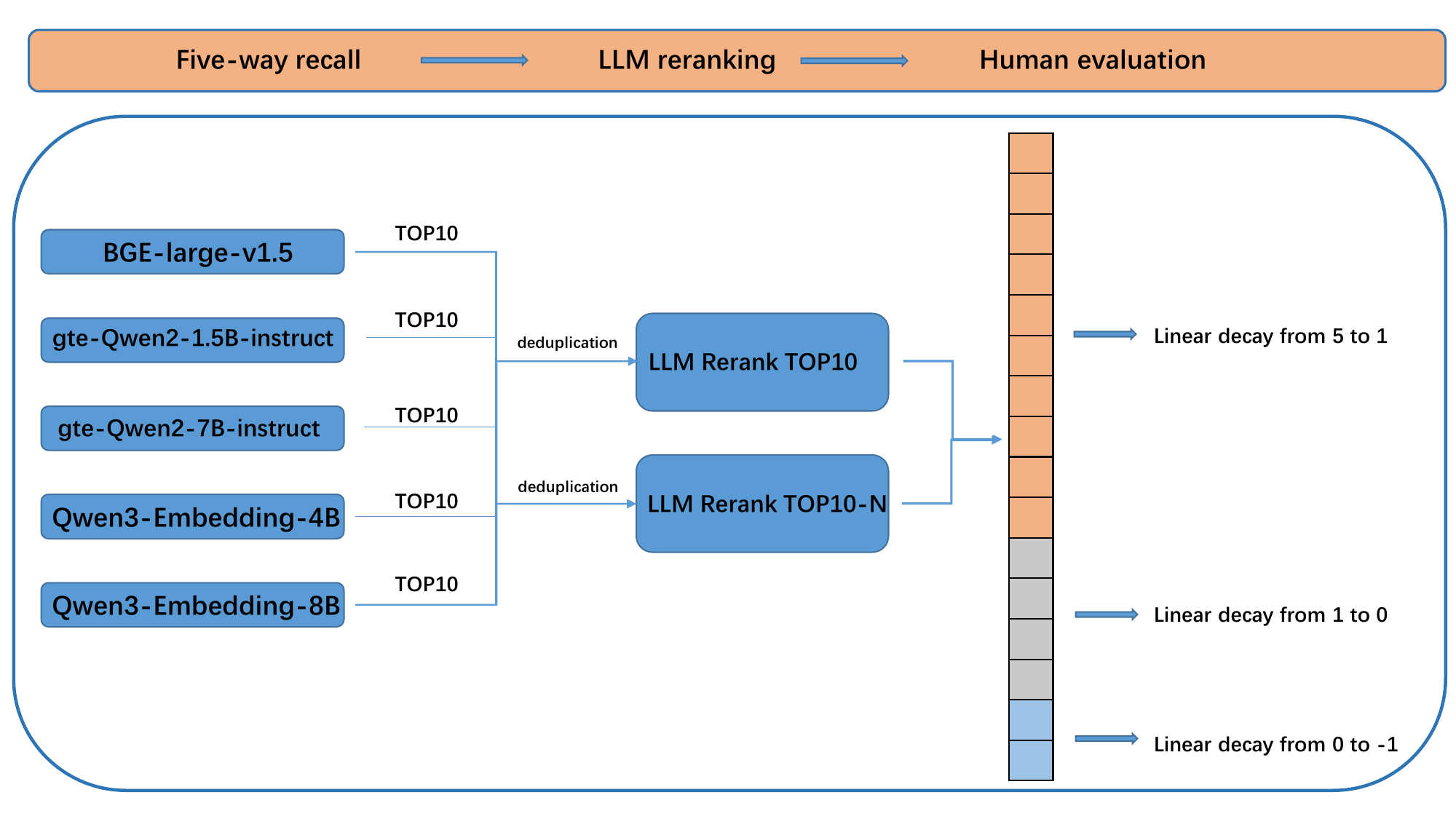}
  \caption{Curated corpus set construction}
  \label{fig:fig8}
\end{figure}
For each question-answer pair, we use BGE-large-v1.5, gte-Qwen2-1.5B-instruct, gte-Qwen2-7B-instruct, Qwen3-Embedding-4B, and Qwen3-Embedding-8B to retrieve the \texttt{TOP10} most similar texts from the corpus based on the question stem in the question-answer dataset. Remove duplicates from the results of the five models in order, ensuring that the number of unique result texts is between 24 and 46. We then use a large model to assist with sorting, as its capability for long-text sorting tasks is slightly limited. Therefore, we divide the deduplicated texts into two parts: \texttt{TOP10} and \texttt{TOP10} to \texttt{TOPN} (N determined by the number of deduplicated texts). These parts are sorted using QWEN-235B-A22B, respectively. Human evaluators then score the results based on their order. For the \texttt{TOP10} texts, scores decrease linearly from 5 to 1. For the \texttt{TOP10} to \texttt{TOPN} texts, the first two-thirds of the rankings score linearly from 1 to 0, while the last third scores linearly from 0 to -1. The texts and their corresponding scores form a curated corpus used for Retrieval tasks. Additionally, the \texttt{TOP10} texts and their scores form a curated re-ranked set used for Reranker-text tasks.

\clearpage

\section{Correlation between models performances and tasks}
\begin{figure}[H]
\centering 
\begin{subfigure}{.48\textwidth}
   \centering\includegraphics[width=\textwidth]
    {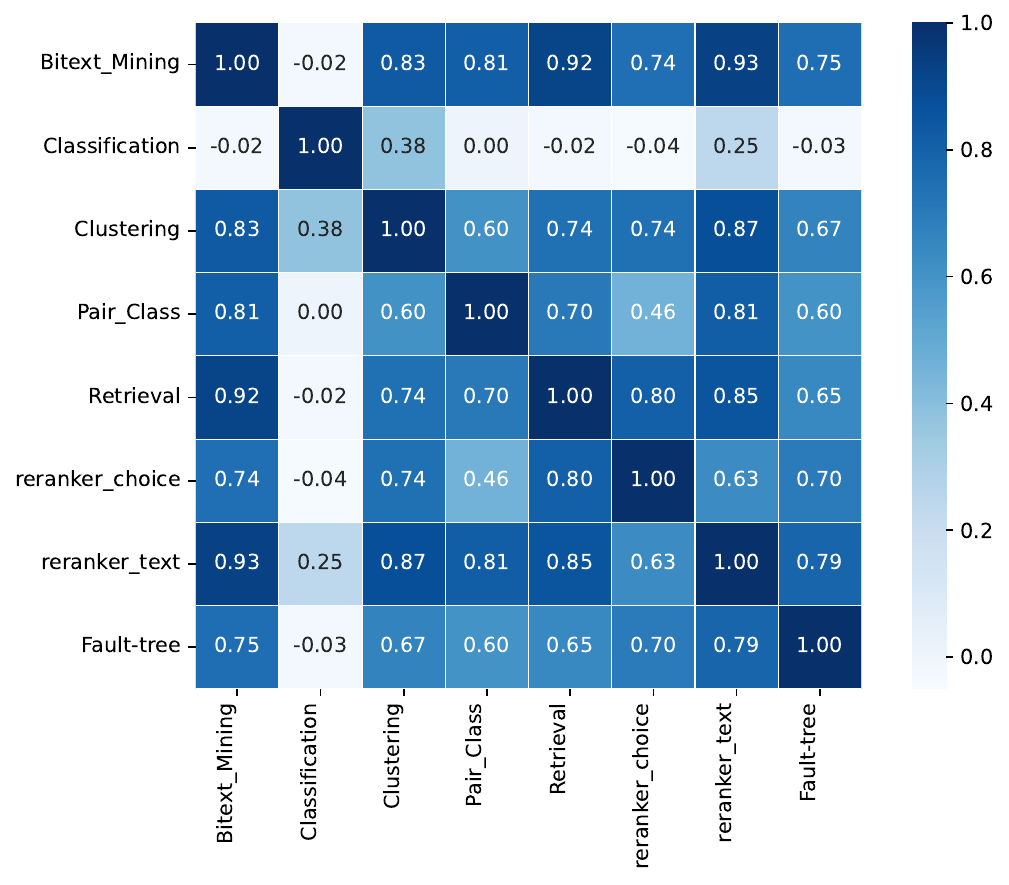} 
   \caption{Tasks Correlation}
\end{subfigure}\hfill%
\begin{subfigure}{.48\textwidth}
   \centering\includegraphics[width=\textwidth]
   {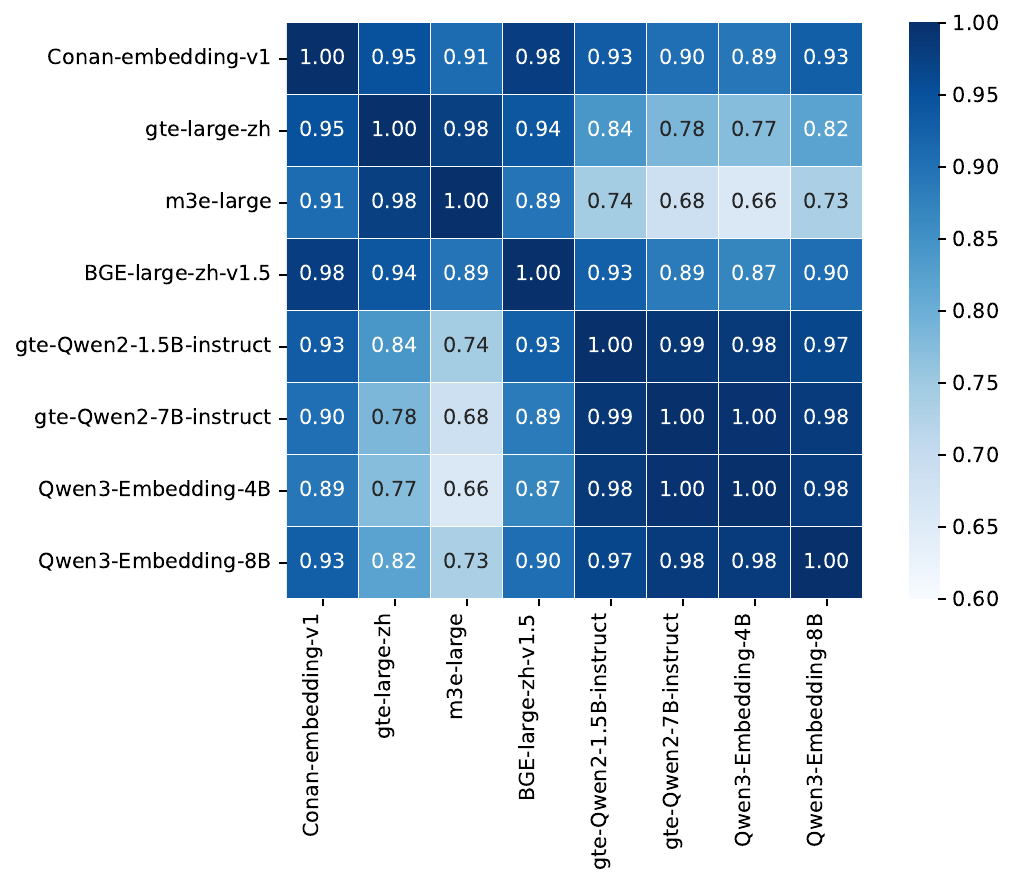} 
   \caption{Models Correlation}
\end{subfigure}
\caption{Correlation between models performances and tasks.}
\label{fig:pair}
\end{figure}

Figure (a) shows the correlation between tasks, where Classification has lower similarity with other tasks. Figure (b) illustrates the correlation between model performances, indicating that m3e-large has lower correlation with other types of models. The gte-Qwen2 series and Qwen3 series have high correlation, which may reflect their closer architectures or similar pre-training tasks.

\section{Reproducing mteb results}
\begin{table}[h]
  \caption{Comparison between official results and reproduced results}
  \centering 
  \small
  \begin{tabular}{lcccccc}
    \toprule
    Task & BitextMining & Classification & Cluster & PairClass & \makecell{Retrieval} & \makecell{Reranker} \\
    \midrule
    Dataset & BUCC.v2(zh-en) & IFlyTek &  \makecell{ThuNews\\ClusteringP2P} &  \makecell{PawsXPair\\Classification(eng)} &  \makecell{Medical\\Retrieval} &  \makecell{CMedQAv2\\reranking} \\
    \midrule
    Source &  Multi-mteb & Cmteb & Cmteb &  Multi-mteb & Cmteb & Cmteb \\
    \midrule
    Conan-embedding-v1 & - / 74.35 & 51.94 / 47.12 & 77.84 / 77.90 & - / 72.59 & 67.94 / 67.99 & 89.72 / 89.74 \\
    gte-large-zh & - / 65.60 & 49.60 / 49.63 & 68.36 / 69.96 & - / 66.45 & 62.88 / 62.88 & 86.46 / 86.46 \\ 
    m3e-large & - / 58.82 & 43.96 / 43.65 & 60.39 / 58.83 & - / 56.33 & 48.66 / 48.66 & 78.27 / 78.27 \\
    BGE-large-zh-v1.5 & - / 87.36 & 48.74 / 48.76 & 59.61 / 59.44 & - / 67.08 & 59.59 / 59.60 & 85.44 / 85.23\\
    gte-Qwen2-1.5B-instruct & 98.80 / 99.30 & 44.85 / 51.75 & 68.24 / 68.64 & 66.03 / 74.82 & 58.65 / 57.92 & 88.12 / 89.12 \\
    gte-Qwen2-7B-instruct & 98.71 / 99.23 & 54.52 / 53.61 & 86.08 / 85.93 & 75.24 / 83.97 & 65.59 / 62.84 & 89.31 / 89.51\\
    Qwen3-Embedding-4B & 98.89 / 99.23 & 54.24 / 54.08 & 87.88 / 88.03 & 68.91 / 76.00 & 63.94 / 63.94 & 85.06 / 86.69\\
    Qwen3-Embedding-8B & 98.88 / 99.23 & 54.74 / 54.81 & 88.79 / 88.33 & 74.59 / 81.08 & 65.47 / 65.73 & 86.39 / 86.83 \\
    \bottomrule 
  \end{tabular}
  \label{tab:table}
\end{table}

We selected tasks and datasets similar to CAMB on mteb for result reproduction, including BitextMining and PairClass which chose the zh-en subset and eng subset. The table shows official results on the left and reproduced results on the right. From the results, it can be seen that the reproduced results are not significantly different from the official results.

\section{Fault-tree evaluation and question difficulty rating}
\subsection{Evaluation}
\begin{figure}[H]
  \centering
  \includegraphics[width=1.0\textwidth]{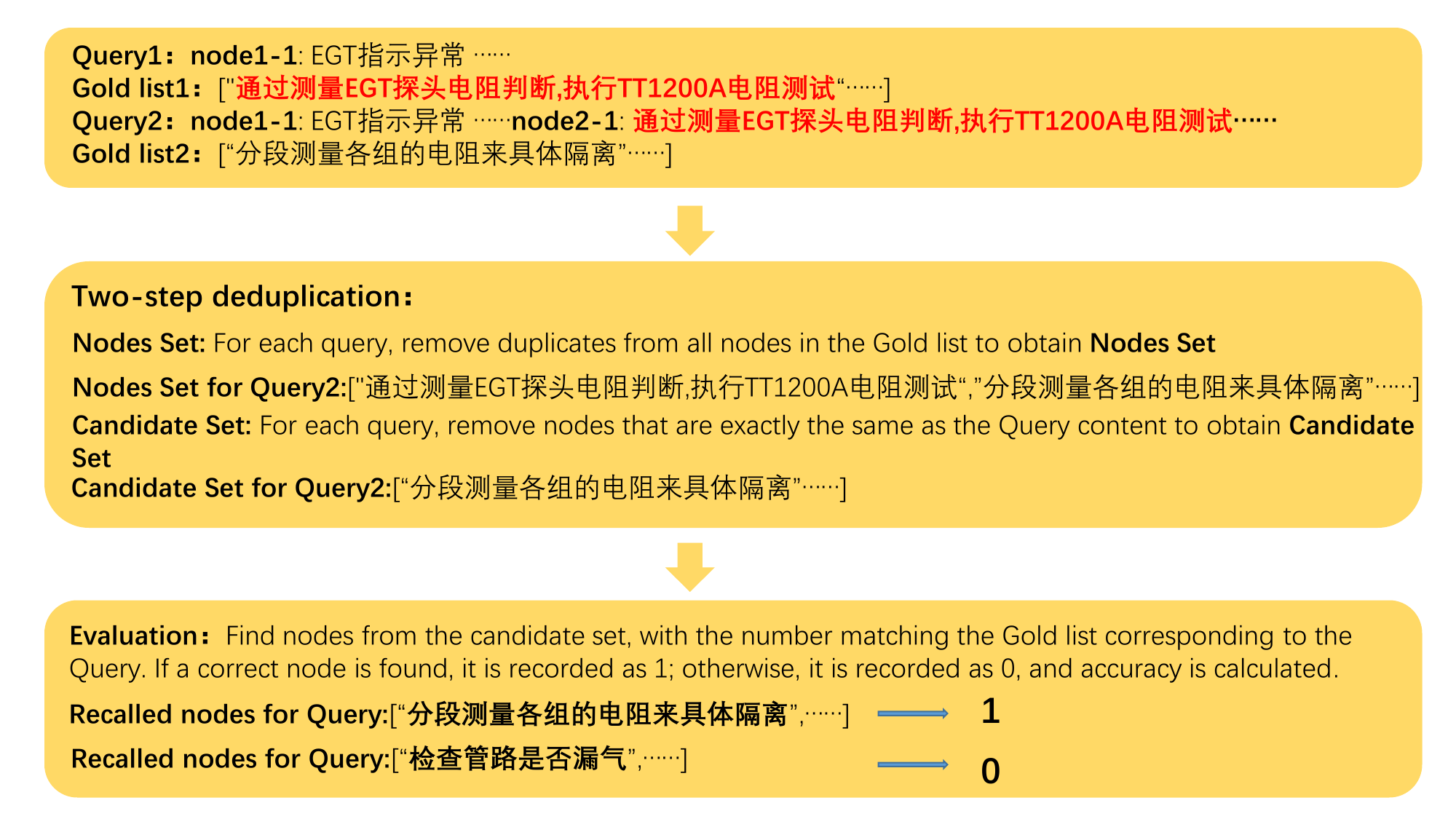}
  \caption{Fault-tree evaluation}
  \label{fig:fig8}
\end{figure}
Due to the issue of complete overlap between the Query and the Gold list corresponding to other Queries in the fault-tree dataset, we adopted a two-step deduplication evaluation method, as shown in Figure 12.

\subsection{Difficulty Rating}
\subsubsection{Complete Failure (All Answers Wrong)}
The probability $P$ of the model answering \textbf{all} nodes incorrectly is used as the difficulty metric. \\
Higher $P$ indicates the model is more likely to fail, thus \textbf{higher difficulty}:
\[ 
\text{Difficulty} \propto \frac{\binom{H-A}{A}}{\binom{H}{A}} \quad \text{(increases with value)}
\]
where:
\begin{itemize}
    \item $H$ = Total number of possible answers 
    \item $A$ = Number of attempted answers 
\end{itemize}
 
\subsubsection{Partial Success (At Least One Correct Answer)} 
The probability $P$ of the model answering \textbf{at least one} node correctly is used as the difficulty metric. \\
Higher $P$ indicates the model is more likely to succeed, thus \textbf{lower difficulty}:
\[ 
\text{Difficulty} \propto \frac{\binom{H-A}{A-C}\binom{A}{C}}{\binom{H}{A}} \quad \text{(decreases with value)}
\]
where:
\begin{itemize}
    \item $C$ = Number of correct answers ($C \geq 1$)
\end{itemize}

 
\bibliography{paper}  

\begin{thebibliography}{28}
\providecommand{\natexlab}[1]{#1}
\providecommand{\url}[1]{\texttt{#1}}
\expandafter\ifx\csname urlstyle\endcsname\relax
  \providecommand{\doi}[1]{doi: #1}\else
  \providecommand{\doi}{doi: \begingroup \urlstyle{rm}\Url}\fi

\bibitem[Albakkoush et~al.(2020)Albakkoush, Pagone, and Salonitis]{pcj_add_2}
Salah Albakkoush, Emanuele Pagone, and Konstantinos Salonitis.
\newblock Scheduling challenges within maintenance repair and overhaul operations in the civil aviation sector.
\newblock 2020.

\bibitem[Dave et~al.(2024)Dave, Nguyen, and Vilim]{lcy1}
Akshay~J. Dave, Tat~Nghia Nguyen, and Richard~B. Vilim.
\newblock Integrating llms for explainable fault diagnosis in complex systems, 2024.
\newblock URL \url{https://arxiv.org/abs/2402.06695}.

\bibitem[Decroos et~al.(2019)Decroos, Bransen, Van~Haaren, and Davis]{pcj1}
Tom Decroos, Lotte Bransen, Jan Van~Haaren, and Jesse Davis.
\newblock A quantitative and qualitative evaluation of llm-based explainable fault localization.
\newblock In \emph{Proceedings of the 25th {ACM SIGKDD} International Conference on Knowledge Discovery and Data Mining}, pages 1851--1861, New York, NY, USA, 2019. ACM.
\newblock \doi{10.1145/3292500.3330758}.

\bibitem[Dong and Kim(2018)]{pcj_add_1}
Ting Dong and Nam~H Kim.
\newblock Cost-effectiveness of structural health monitoring in fuselage maintenance of the civil aviation industry.
\newblock \emph{Aerospace}, 5\penalty0 (3):\penalty0 87, 2018.

\bibitem[Hadash et~al.(2018)Hadash, Kermany, Carmeli, Lavi, Kour, and Jacovi]{pcj4}
Guy Hadash, Einat Kermany, Boaz Carmeli, Ofer Lavi, George Kour, and Alon Jacovi.
\newblock Development of an aerospace engineering evaluation set for large language model benchmarking.
\newblock \emph{arXiv preprint arXiv:1804.09028}, 2018.

\bibitem[Heesch et~al.(2024)Heesch, Eilermann, Windmann, Diedrich, Rosenthal, and Niggemann]{pcj2}
Ren{\'e} Heesch, Sebastian Eilermann, Alexander Windmann, Alexander Diedrich, Philipp Rosenthal, and Oliver Niggemann.
\newblock Evaluating large language models for real-world engineering tasks.
\newblock In \emph{2014 14th International Conference on Frontiers in Handwriting Recognition (ICFHR)}. IEEE, 2024.

\bibitem[Hou et~al.(2025)Hou, Jia, Xing, Chen, and Du]{pcj_add_4}
Lei Hou, Beixi Jia, Chenguang Xing, Zhaojiang Chen, and Ziliang Du.
\newblock Applied research on an aircraft maintenance assistant based on a large language model.
\newblock In \emph{Proceedings of the 2025 4th International Conference on Intelligent Systems, Communications and Computer Networks}, ISCCN '25, page 1–7, New York, NY, USA, 2025. Association for Computing Machinery.
\newblock ISBN 9798400715204.
\newblock \doi{10.1145/3732945.3732946}.
\newblock URL \url{https://doi.org/10.1145/3732945.3732946}.

\bibitem[Kour and Saabne(2014)]{pcj3}
George Kour and Raid Saabne.
\newblock A multi-agent approach to fault localization via graph-based retrieval and reflexion.
\newblock In \emph{Soft Computing and Pattern Recognition (SoCPaR), 2014 6th International Conference of}, pages 312--318. IEEE, 2014.

\bibitem[Laskar et~al.(2024)Laskar, Alqahtani, Bari, Rahman, Khan, Khan, Jahan, Bhuiyan, Tan, Parvez, Hoque, Joty, and Huang]{zyh3}
Md~Tahmid~Rahman Laskar, Sawsan Alqahtani, M~Saiful Bari, Mizanur Rahman, Mohammad Abdullah~Matin Khan, Haidar Khan, Israt Jahan, Amran Bhuiyan, Chee~Wei Tan, Md~Rizwan Parvez, Enamul Hoque, Shafiq Joty, and Jimmy Huang.
\newblock A systematic survey and critical review on evaluating large language models: Challenges, limitations, and recommendations.
\newblock In \emph{Proceedings of the 2024 Conference on Empirical Methods in Natural Language Processing (EMNLP)}, pages 13785--13816, Miami, Florida, USA, November 2024. Association for Computational Linguistics.
\newblock \doi{10.18653/v1/2024.emnlp-main.764}.

\bibitem[Lin et~al.(2025)Lin, Zhang, Fu, and Liu]{zyh2}
Lin Lin, Sihao Zhang, Song Fu, and Yikun Liu.
\newblock {FD-LLM}: Large language model for fault diagnosis of machines.
\newblock \emph{Advanced Engineering Informatics}, 65, Part A, 2025.
\newblock \doi{https://doi.org/10.1016/j.aei.2025.103208}.

\bibitem[Liu et~al.(2025)Liu, Cui, Hu, Li, Lin, and Zhang]{lcy5}
Beiming Liu, Zhizhuo Cui, Siteng Hu, Xiaohua Li, Haifeng Lin, and Zhengxin Zhang.
\newblock Llm evaluation based on aerospace manufacturing expertise: Automated generation and multi-model question answering, 2025.
\newblock URL \url{https://arxiv.org/abs/2501.17183}.

\bibitem[LIU et~al.(2024)LIU, Qian, Zhao, and Tao]{pcj6}
Peifeng LIU, Lu~Qian, Xingwei Zhao, and Bo~Tao.
\newblock Joint knowledge graph and large language model for fault diagnosis and its application in aviation assembly.
\newblock \emph{IEEE Transactions on Industrial Informatics}, 20\penalty0 (6):\penalty0 8160--8169, 2024.
\newblock \doi{10.1109/TII.2024.3366977}.

\bibitem[Mangortey et~al.(2025)Mangortey, Singh, Chen, and Sarkhel]{pcj_add_10}
Eugene Mangortey, Satyen Singh, Shuo Chen, and Kunal Sarkhel.
\newblock Aviation language understanding evaluation (alue)--large language model benchmark with aviation datasets.
\newblock In \emph{AIAA AVIATION FORUM AND ASCEND 2025}, page 3247, 2025.

\bibitem[Meng et~al.(2025)Meng, Jiao, Li, Wang, Pan, Jing, and Tang]{lcy3}
Xiangzhen Meng, Xiaoxuan Jiao, Jiahui Li, Shenglong Wang, Jinxin Pan, Bo~Jing, and Xilang Tang.
\newblock Cellmea:a collaboratively enhanced large language model-based entity alignment for aircraft fault maintenance.
\newblock \emph{Expert Systems with Applications}, 282:\penalty0 127630, 2025.
\newblock ISSN 0957-4174.
\newblock \doi{https://doi.org/10.1016/j.eswa.2025.127630}.
\newblock URL \url{https://www.sciencedirect.com/science/article/pii/S0957417425012527}.

\bibitem[Oderinde et~al.(2025)Oderinde, Chandra, Albertoli, Bhanpato, Bendarkar, and Mavris]{pcj_add_11}
Timilehin~P Oderinde, Chetan Chandra, Leslie Albertoli, Jirat Bhanpato, Mayank~V Bendarkar, and Dimitri Mavris.
\newblock Aviation safety qa dataset for extracting knowledge from incident reports.
\newblock In \emph{AIAA AVIATION FORUM AND ASCEND 2025}, page 3248, 2025.

\bibitem[Qiao et~al.(2024)Qiao, Tian, Xu, Yang, Yang, and Wang]{pcj_add_12}
Dongxiao Qiao, Xianhui Tian, Yubin Xu, Jie Yang, Le~Yang, and Xuhui Wang.
\newblock Cast-eval: A domain-specific benchmark for large language models in civil aviation safety.
\newblock In \emph{2024 IEEE 2nd International Conference on Electrical, Automation and Computer Engineering (ICEACE)}, pages 341--344. IEEE, 2024.

\bibitem[Rosales~Cabezas et~al.(2024)Rosales~Cabezas, Traslavina~Navarrete, et~al.]{pcj_add_8}
Alec~Mauricio Rosales~Cabezas, Danny~Stevens Traslavina~Navarrete, et~al.
\newblock Implementation of a large language model for the interpretation of the colombian aeronautical regulations.
\newblock 2024.

\bibitem[Siddeshwar et~al.(2024)Siddeshwar, Azim, Alwidian, and Makrehchi]{pcj_add_5}
Vaishali Siddeshwar, Akramul Azim, Sanaa Alwidian, and Masoud Makrehchi.
\newblock Towards enhancing aviation safety through advanced incident analysis using large language models.
\newblock In \emph{2024 34th International Conference on Collaborative Advances in Software and COmputiNg (CASCON)}, pages 1--7, 2024.
\newblock \doi{10.1109/CASCON62161.2024.10837802}.

\bibitem[Sobester et~al.(2025)Sobester, Middleton, Yong, Marsh, and Silva]{lcy4}
A~Sobester, S~Middleton, H~Yong, R~Marsh, and E~A Da~Cruz Silva.
\newblock Retrieval-augmented generation and in-context prompted large language models in aircraft engineering.
\newblock In \emph{AIAA SCITECH 2025 Forum}. American Institute of Aeronautics and Astronautics, January 2025.
\newblock \doi{https://doi.org/10.2514/6.2025-0700}.

\bibitem[Syed et~al.(2024)Syed, Light, Guo, Zhang, Qin, Ouyang, and Hu]{pcj_add_9}
Usman Syed, Ethan Light, Xingang Guo, Huan Zhang, Lianhui Qin, Yanfeng Ouyang, and Bin Hu.
\newblock Benchmarking the capabilities of large language models in transportation system engineering: Accuracy, consistency, and reasoning behaviors.
\newblock 08 2024.
\newblock \doi{10.48550/arXiv.2408.08302}.

\bibitem[Tian et~al.(2024)Tian, Hou, Wu, Shu, Liu, Xiang, Gu, Filla, Li, Liu, Chen, Tang, Liu, and Wang]{lcy6}
Jie Tian, Jixin Hou, Zihao Wu, Peng Shu, Zhengliang Liu, Yujie Xiang, Beikang Gu, Nicholas Filla, Yiwei Li, Ning Liu, Xianyan Chen, Keke Tang, Tianming Liu, and Xianqiao Wang.
\newblock Assessing large language models in mechanical engineering education: A study on mechanics-focused conceptual understanding, 2024.
\newblock URL \url{https://arxiv.org/abs/2401.12983}.

\bibitem[Vidyaratne et~al.(2024)Vidyaratne, Lee, Kumar, Watanabe, Farahat, and Gupta]{pcj5}
Lasitha Vidyaratne, Xian~Yeow Lee, Aman Kumar, Tsubasa Watanabe, Ahmed Farahat, and Chetan Gupta.
\newblock Generating troubleshooting trees for industrial equipment using large language models (llm).
\newblock In \emph{2024 IEEE International Conference on Prognostics and Health Management (ICPHM)}, pages 116--125, 2024.
\newblock \doi{10.1109/ICPHM61352.2024.10626823}.

\bibitem[Wan et~al.(2025)Wan, Shen, Li, Sun, Li, and Zhang]{pcj_add_6}
Jia'ang Wan, Feng Shen, Fujuan Li, Yanjin Sun, Yan Li, and Shiwen Zhang.
\newblock Aviationllm: An llm-based knowledge system for aviation training.
\newblock \emph{arXiv preprint arXiv:2506.14336}, 2025.

\bibitem[Wu et~al.(2024)Wu, Velazco, Zhao, Luján, Movva, Roy, Nguyen, Rodriguez, Wu, Albada, Kiseleva, and Mudgerikar]{zyh1}
Yiran Wu, Mauricio Velazco, Andrew Zhao, Manuel Raúl~Meléndez Luján, Srisuma Movva, Yogesh~K Roy, Quang Nguyen, Roberto Rodriguez, Qingyun Wu, Michael Albada, Julia Kiseleva, and Anand Mudgerikar.
\newblock Excytin-bench: Evaluating llm agents on cyber threat investigation.
\newblock \emph{arXiv preprint arXiv:2407.04069}, 2024.

\bibitem[Xie et~al.(2025)Xie, Tang, Gu, and Cui]{zyh4}
Xiaoyue Xie, Xilang Tang, Siwei Gu, and Lijie Cui.
\newblock An intelligent guided troubleshooting method for aircraft based on {HybirdRAG}.
\newblock \emph{Nature}, 15\penalty0 (17752), 2025.

\bibitem[Yang et~al.(2024)Yang, Xiang, and Chen]{pcj_add_7}
Jianzhong Yang, Xinyu Xiang, and Xiyuan Chen.
\newblock A retrieval-augmented generation-based method for aviation accident data analysis.
\newblock In \emph{2024 4th International Conference on Artificial Intelligence, Robotics, and Communication (ICAIRC)}, pages 868--874, 2024.
\newblock \doi{10.1109/ICAIRC64177.2024.10900039}.

\bibitem[Zheng et~al.(2024)Zheng, Pan, Liu, and Chen]{lcy2}
Shuwen Zheng, Kai Pan, Jie Liu, and Yunxia Chen.
\newblock Empirical study on fine-tuning pre-trained large language models for fault diagnosis of complex systems.
\newblock \emph{Reliability Engineering \& System Safety}, 252:\penalty0 110382, 2024.
\newblock ISSN 0951-8320.
\newblock \doi{https://doi.org/10.1016/j.ress.2024.110382}.
\newblock URL \url{https://www.sciencedirect.com/science/article/pii/S095183202400454X}.

\bibitem[ZIO et~al.(2019)ZIO, FAN, ZENG, and KANG]{pcj_add_3}
Enrico ZIO, Mengfei FAN, Zhiguo ZENG, and Rui KANG.
\newblock Application of reliability technologies in civil aviation: Lessons learnt and perspectives.
\newblock \emph{Chinese Journal of Aeronautics}, 32\penalty0 (1):\penalty0 143--158, 2019.
\newblock ISSN 1000-9361.
\newblock \doi{https://doi.org/10.1016/j.cja.2018.05.014}.
\newblock URL \url{https://www.sciencedirect.com/science/article/pii/S1000936118301948}.

\end{thebibliography}

\end{document}